%% file: ms.tex
\documentclass[11pt,a4paper]{article}
\usepackage[hyperref]{emnlp2020}
\usepackage{times}
\usepackage{latexsym}
\usepackage{graphicx}
\usepackage{amsmath}
\usepackage{caption}
\usepackage{subcaption}

\aclfinalcopy

\title{On the Language-specificity of Multilingual BERT\\and the Impact of Fine-tuning}
\author{
    Marc Tanti$^1$ \ \ ~~Lonneke van der Plas$^2$ \ \ ~~Claudia Borg$^3$ \ \ ~~Albert Gatt$^4$\\\\
    $^1$University of Malta, Institute of Linguistics and Language Technology \\
    $^2$Idiap Research Institute \\
    $^3$University of Malta, Department of AI \\
    $^4$Utrecht University, Information and Computing Sciences\\
    \texttt{\{marc.tanti,claudia.borg\}@um.edu.mt}\\\texttt{lonneke.vanderplas@idiap.ch, a.gatt@uu.nl}
}

\begin{document}

\maketitle
\begin{abstract}
    Recent work has shown evidence that the knowledge acquired by multilingual BERT (mBERT) has two components: a language-specific and a language-neutral one.
    This paper analyses the relationship between them, in the context of fine-tuning on two tasks -- POS tagging and natural language inference -- which require the model to bring to bear different degrees of language-specific knowledge.
    Visualisations reveal that mBERT loses the ability to cluster representations by language after fine-tuning, a result that is supported by evidence from language identification experiments.
    However, further experiments on `unlearning' language-specific representations using gradient reversal and iterative adversarial learning are shown not to add further improvement to the language-independent component over and above the effect of fine-tuning.
    The results presented here suggest that the process of fine-tuning causes a reorganisation of the model's limited representational capacity, enhancing language-independent representations at the expense of language-specific ones.
\end{abstract}

\section{Introduction}

Since the introduction of transformer architectures and the demonstration that they improve the state of the art on tasks such as machine translation and parsing \cite{Vaswani2017}, there has been a decisive turn in NLP towards the development of large, pre-trained transformer models, such as BERT \cite{devlin-etal-2019-bert}.
Such models are pre-trained on tasks such as masked language modelling (MLM) and next-sentence prediction (NSP) and are intended to be task-agnostic, facilitating their transfer to new tasks following fine-tuning with limited amounts of data.
 
Extending such models to multiple languages is a natural next step, as evidenced by the recent proliferation of multilingual transformers, including multilingual BERT (mBERT), XLM \cite{Conneau2019}, and XLM-R \cite{Conneau2020}.
These follow from a line of earlier work which sought to achieve transferable multilingual representations using recurrent network-based methods \cite[e.g.][{\em inter alia}]{Artetxe2019}, as well as work on developing multilingual embedding representations \cite{Ruder2017}.

The considerable capacity of these multilingual models and their success in cross-lingual tasks has motivated a lot of research into the nature of the representations learned during pre-training.
On the one hand, there is a significant amount of research suggesting that models such as mBERT acquire robust language-specific representations \cite{wu-dredze-2019-beto,libovicky-etal-2020-language,ChoenniWhat2020}.
On the other hand, it has been suggested that in addition to language-specific information, models like mBERT also have language-neutral representations, which cut across linguistic distinctions and enable the model to handle aspects of meaning language-independently.
This also allows the model to be fine-tuned on a monolingual labelled data set and achieve good results in other languages, a process known as cross-lingual zero-shot learning
\cite{pires-etal-2019-multilingual, libovicky-etal-2020-language, conneau-etal-2018-xnli, Hu2020}.
These results have motivated researchers to try and disentangle the language-specific and language-neutral components of mBERT \cite[e.g.][]{libovicky-etal-2020-language,gonen-etal-2020-greek}.

This background provides the motivation for the work presented in this paper.
We focus on the relationship between language-specific and language-neutral representations in mBERT.
However, our main goal is to study the impact of fine-tuning on the balance between these two types of representations.
More specifically, we measure the effect of fine-tuning on mBERT's representations in the context of two different tasks -- part-of-speech (POS) tagging and natural language inference (NLI) -- which lay different demands on the model's semantic and language-specific knowledge.
While NLI involves reasoning about deep semantic relations between texts, POS tagging requires a model to bring to bear knowledge of a language's morphosyntactic features.
Though many languages share such features as a result of typological relations \cite[which mBERT is known to exploit; see, e.g.][]{pires-etal-2019-multilingual,ChoenniWhat2020,rama-etal-2020-probing}, there are also language-specific features to which, we hypothesise, mBERT needs to dedicate a greater share of its representational capacity, compared to the NLI task.

We show that the model accommodates language-specific and language-neutral representations to different degrees as a function of the task it is fine-tuned on.
This is supported by results from language identification (LID) experiments, conducted both on task-specific data and on a new data set extracted from Wikipedia.
We then consider two alternative strategies that force the model to `unlearn' language-specific representations, via gradient reversal or iterative adversarial learning.
These are shown not to further improve the language-independent component for cross-lingual transfer, over and above the effect of fine-tuning.
Thus, we conclude that the reorganisation of mBERT's representations that happens with fine-tuning is already taking on this role.
Note that our goal is not to improve mBERT's multilinguality but to acquire a better understanding of it, extending previous work along these lines.

Our main contributions are (a) to provide further support for the distinction between language-specific and language-neutral representation in mBERT; (b) to show that fine-tuning results in a reorganisation of mBERT's representations in a way that destroys existing language clusters; (c) to study two methods to enhance language-neutrality in mBERT, both of which are shown not to improve performance on fine-tuned tasks; (d) a  new Wikipedia-based language identification data set.

\section{Tasks and data}

To perform cross-lingual zero-shot learning, we fine-tune mBERT on English only and evaluate it on multiple languages at once.
We focus on two tasks: the low-level structured prediction task of cross-lingual POS tagging and the high-level semantic task of  cross-lingual NLI.
We chose these two tasks, because each task requires a different type of linguistic knowledge.
Cross-lingual POS tagging requires a model to bring to bear language-specific knowledge related to morphosyntactic properties, word order, etc., in determining the correct part-of-speech to assign to a token in context.
Previous work, for example by \citet{pires-etal-2019-multilingual}, showed good results on this task for mBERT in a cross-lingual zero-shot setting.
In contrast, NLI requires a model to determine the semantic relationship between two texts, determining whether it is one of entailment, contradiction or neutrality \cite{Sammons2015,Bowman2015}.
We expected this task to require more language-neutral, semantic knowledge, compared to POS tagging.\footnote{
    We note that one task is word-level and one is sentence-level.
    Ideally the two tasks would have the same level of granularity, but, to our knowledge, no two tasks conform to this goal while at the same time addressing the morphosyntax-versus-semantics requirements.
}

In addition to results on these two tasks, we report results on language identification (LID) experiments.
These are reported both on the test data for the tasks themselves, as well as on an independent data set consisting of Wikipedia texts, described below.

In all the experiments reported, we reserve a development set for hyperparameter tuning and a validation set to monitor progress during training.
Data set statistics are provided in the Appendix.
Following the practice of \citet{Hu2020} for the XTREME benchmark, all texts were truncated to their first 128 tokens (with XNLI having a combined premise-hypothesis length of 128).
All the data was tokenised using the {\tt bert-base-multilingual-cased} tokeniser.\footnote{
    \url{https://huggingface.co/bert-base-multilingual-cased}
}
Data containing unknown tokens (according to the tokeniser) was omitted.
All our experiments are conducted on data for 33 languages, the same set of languages included in the UDPOS task of the XTREME benchmark \cite{Hu2020}.\footnote{
    ISO 639-1 language codes used: af, ar, bg, de, el, en, es, et, eu, fa, fi, fr, he, hi, hu, id, it, ja, kk, ko, mr, nl, pt, ru, ta, te, th, tl, tr, ur, vi, yo, and zh.
}
The exception is XNLI, for which cross-lingual test data exists for only 15 of these languages \cite{conneau-etal-2018-xnli}.

\paragraph{UDPOS} For POS tagging, we use data from the Universal Dependencies Treebank \cite[UDPOS;][]{Marneffe2020} v2.7, using the train/dev/test splits provided.
A validation set is randomly sampled from the training set.
Since we are interested in cross-lingual zero-shot learning, we removed all non-English data from the train/val splits.

\paragraph{XNLI} For NLI, we use the monolingual English MultiNLI data set \cite{williams-etal-2018-broad} as a training set, and the Cross-lingual Natural Language Inference data \cite[XNLI;][]{conneau-etal-2018-xnli} for the development set and test set.
Again, a validation set is randomly sampled from the training set.

\paragraph{Wikipedia} An independent data set for LID was extracted from Wikipedia.
For each language, we randomly selected 5\,000 paragraphs which are at least 100 characters in length.
Further details are provided in the Appendix.

\section{Model architecture}

\begin{figure}[!t]
    \centering
	\includegraphics[width=4cm]{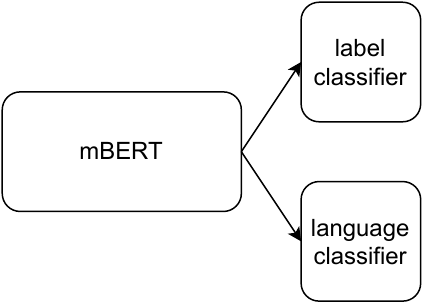}
	\caption{The basic model architecture.}
	\label{fig:architecture}
\end{figure}

The general architecture used in our experiments is shown in Figure~\ref{fig:architecture}.
We use a pre-trained mBERT model to encode the input using its final hidden layer, because we are interested in the linguistic capabilities of mBERT's typical use case, which is also the practice in the XTREME benchmark \cite{Hu2020}.
The same mBERT model is shared between two single-layer softmax classifiers, both trained using a categorical cross-entropy loss.
One of these assigns a task-specific label (a POS tag or an NLI class).
This is trained on UDPOS or XNLI data.
We will sometimes refer to this as the {\em target} task.
The other is a language classifier that predicts which of the 33 languages (see above) the input text is written in.
The language classifier is trained on the Wikipedia data.

We conduct separate experiments for each target task (UDPOS and XNLI).
In the case of UDPOS, the classification is for individual tokens.
For XNLI, the classification is for sentence pairs, represented as a single text consisting of the concatenation of the premise and hypothesis, separated by a \texttt{[SEP]} token.

The language classifier is also trained to either predict the language of each token or of an entire text according to the target task.
In this way, we are also able to test this classifier for predictions both on the independent Wikipedia data and on the test data from the target task (which is also labelled by language).

Unless otherwise specified, gradients from the language classifier are not propagated to the pre-trained mBERT model.
Thus, the mBERT model parameters are only fine-tuned on the target task data set whilst the language classifier is fine-tuned in isolation.
This allows us to monitor how much language-specific information exists in the mBERT encoding, without directly influencing it.

More information about how the model was trained together with the hyperparameters used can be found in the Appendix.\footnote{
    Our implementation can be found on the GitHub repo \url{https://github.com/mtanti/mbert-language-specificity}.
}

\subsection{Modifying mBERT's language-specific representations}

To explain the rationale for our experiments on language-specificity, we distinguish {\em language-neutrality} from {\em language confusion}.

Let $T_{L1}$ be a text in language L1, and its translation $T_{L2}$ in language L2, and let $E(\cdot)$ be an encoding extracted from a model $\mathcal{M}$.
We say that $E(\cdot)$ is language-neutral if $E(T_{L1})$ is close to, or identical with, $E(T_{L2})$.
Thus, $\mathcal{M}$ treats semantically equivalent texts in different languages in the same way.
In contrast, language confusion places a weaker requirement on a model: here, $E(T_{L1})$ and $E(T_{L2})$ need not be identical, but the encoding itself does not support language identification.
In short, language confusion arises when language-specific information is missing from an encoding.

Here, we focus on reducing language-specificity using methods to enhance language confusion, inspired by work on domain adaptation.
In a domain adaptation setting, \citet{Ganin2015} successfully enhanced domain invariance by enhancing domain confusion.
We want to see if this works for language as well, that is, whether reducing the language-specificity of representations leads to better cross-lingual generalisation on target tasks.
To this end, we consider two methods which are intended to enhance the model's language confusion: gradient reversal and iterative entropy maximisation.\footnote{
    Note that we do not change the architecture or the pre-trained mBERT model in these experiments.
}

\paragraph{Gradient reversal} Gradient reversal has been used to train a classifier in a multi-domain setting, using only labelled data in a single domain \cite{Ganin2015}.
A small labelled data set in one domain and a large unlabelled data set in another domain can be used to encourage the features learned by the model to be domain-invariant by training the model to confuse a domain classifier, thus avoiding any features that are specific to the labelled data set's domain.
We make use of \citeauthor{Ganin2015}'s gradient reversal layer, which multiplies gradients from the language classifier by a factor $-\lambda$ before backpropagating them to the mBERT model.
In effect, this updates the language classifier layer itself to perform better at LID whilst the mBERT parameters are updated to make the language of an encoded text progressively harder to classify.
A similar strategy was proposed by \citet{libovicky-etal-2020-language}, who used gradient reversal from LID during additional pre-training of mBERT using masked language modelling.
Here, we consider its use in the context of fine-tuning on a target task, such as UDPOS or NLI.
Hyperparameters, including $\lambda$, are listed in the Appendix.

\paragraph{Entropy maximisation} Iterative entropy maximisation is closer in spirit to the strategies employed by generative adversarial networks (GANs), in that training is performed iteratively in two alternating phases, each lasting one epoch.
In the first phase, the language classification layer is trained in isolation for one epoch on the Wikipedia data set.
In the second phase, the target classifier is trained together with mBERT with two goals: (a) maximise the accuracy of the label classifier and (b) maximise the entropy of the language classifier's output probabilities.
The entropy maximisation step is intended to make the language classifier approach a uniform distribution, thus training the mBERT model to make the encodings as confusing as possible to the language classifier.

The loss in the second phase is a combination of the label classifier's loss and the language classifier's negative entropy.
This loss is calculated as follows:
\begin{equation}
	L = (1-w)(\text{XE}(y_a)) + w\left(-\sum\ln{y_b}\right)
\end{equation}
\noindent
where $y_a$ is the label classifier output, $y_b$ is the language classifier output, XE is the cross entropy function, and $w$ is the weight assigned to the loss for the language classifier.
Hyperparameters, including $w$, are listed in the Appendix.

\section{Experiments}

In what follows, we first discuss the language clustering and LID capabilities of mBERT both in its unmodified form, and after fine-tuning on UDPOS and XNLI.
Then, we consider the impact of gradient reversal and entropy maximisation, both of which seek to shift mBERT representations towards greater language neutrality.

\subsection{The effect of fine-tuning on language-specific representations}

In this section we focus on how much language-specific information is readily available in mBERT representations before and after fine-tuning it.
We measure this in two steps.

First, we train the label and language classifiers on the unmodified pre-trained mBERT model, without modifying mBERT's parameters.
We measure the model's performance on the target task using the label classifier and on the LID task using the language classifier.
We measure LID performance on both the target task test data and on the Wikipedia test data.

Second, we reinitialise both classifiers and train the label classifier together with the pre-trained mBERT model, after which we freeze mBERT's parameters and train the language classifier.
Again, we measure the label and language classifiers' performance as explained above.

In addition, we show 2D t-SNE projections of mBERT's representations of a sample of data items (tokens or texts) from the target task test set and from the Wikipedia test set.
Sample sizes are given in the Appendix.
We colour the 2D points according to label (target test set) or language (target test set and Wikipedia test set).
This allows us to compare the way in which mBERT organises representations prior to and after fine-tuning.
As a numerical measure of organisation we cluster the full representations (prior to t-SNE compression) using k-means and measure the agreement of the clusters for labels/languages using the V-measure \cite{Rosenberg2007}.
We take the average V-measure from ten independent runs of k-means clustering (from scratch).

Table~\ref{tab:finetuning_results} gives the macro F1 scores of the classifier predictions.
For both target tasks, performance improves after fine-tuning, as expected.
This occurs to a greater extent on XNLI than on UDPOS.\footnote{
    Note that we report macro F1 scores.
    The results reported in related work, such as the XTREME benchmark \cite{Hu2020}, report accuracies for NLI and micro F1 scores for POS.
    They are comparable to those in Table~\ref{tab:finetuning_results}: we obtain micro F1 for POS of 73.4\%, compared to 70.3\% reported by \citet{Hu2020}; and accuracy of 66.3\% on XNLI, where the equivalent result reported by \citet{Hu2020} is 65.4\%.
}
For LID, we observe the reverse trend: on both the target test sets and Wikipedia, the classifier's performance in detecting the language of the input drops significantly after fine-tuning for UDPOS, less so for XNLI (with Wikipedia LID remaining unchanged on XNLI).

\begin{table}[!t]
    \centering
    \small
    \begin{tabular}{l|cc|cc} 
        \hline
        & \multicolumn{2}{c|}{\bf UDPOS} & \multicolumn{2}{c}{\bf XNLI} \\
        \hline
        & Init. & Fine-T. & Init. & Fine-T. \\
        \hline
        {\bf Target task} &  51.2 & 59.6 & 29.7 & 66.3 \\
        \hline
        {\bf Lang. ID (Target)} & 78.3 & 0.3 & 49.8 & 39.2 \\
        {\bf Lang. ID (Wiki)} & 59.3 & 0.5 & 97.0 & 97.2 \\
        \hline
    \end{tabular}
    \caption{
        Macro F1 scores (\%) for target tasks (UDPOS and XNLI) and language identification before (Init.) and after fine-tuning (Fine-T.).
        Note that `Lang. ID (Target)' refers to language classification on the target data set.
    }
    \label{tab:finetuning_results}
\end{table}

\begin{table}[h]
	\centering
	\small
	\begin{tabular}{l|cc|cc} 
		\hline
		& \multicolumn{2}{c|}{\bf UDPOS} & \multicolumn{2}{c}{\bf XNLI} \\
		\hline
		& Grad. & Ent. & Grad. & Ent. \\
		\hline
		{\bf Target task} &  53.5 & 56.8 & 62.2 & 62.1 \\
		\hline
		{\bf Lang. ID (Target)} & 0.1 & 5.5 & 1.3 & 3.4 \\
		{\bf Lang. ID (Wiki)} & 0.1 & 3.1 & 1.5 & 54.3 \\
		\hline
	\end{tabular}
	\caption{
		Macro F1 scores (\%) for target tasks (UDPOS and XNLI) and language identification after training using gradient reversal (Grad.) and entropy maximisation (Ent.).
		Note that `Lang. ID (Target)' refers to language classification on the target data set.
	}
	\label{tab:langunlearning_results}
\end{table}

Figure~\ref{fig:finetuning_udpos_plots} and Figure~\ref{fig:finetuning_xnli_plots} show the t-SNE projections of the token-based UDPOS representations and the text-based XNLI representations respectively, together with the corresponding V-measure.
For labels in both target tasks, mBERT starts off with no discernible structure, whereas fine-tuning results in clear clusters by label (compare Figures \ref{fig:finetuning_udpos_frozen_label_label} vs \ref{fig:finetuning_udpos_finetuned_label_label}; and \ref{fig:finetuning_xnli_frozen_label_label} vs \ref{fig:finetuning_xnli_finetuned_label_label}).\footnote{
    Note that fine-tuning was done on English data, while the plots are generated on the multilingual test data, meaning that data for languages that were not used for fine-tuning nevertheless cluster by labels in line with the English data.
}

\begin{figure*}
    \centering
	\begin{subfigure}{6cm}
	    \centering
		\includegraphics[width=6cm]{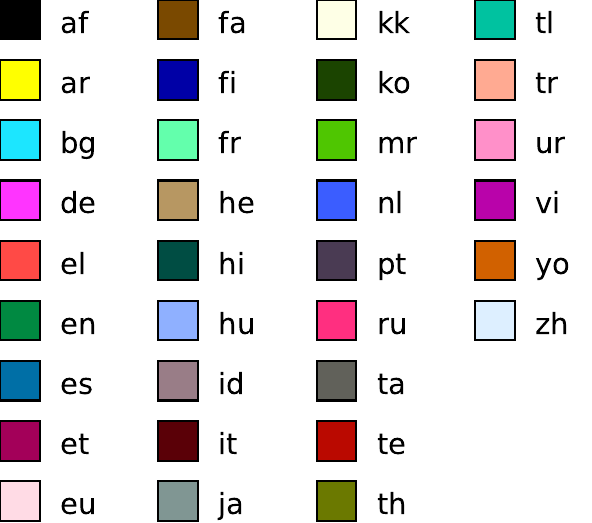}
		\caption{Colour legend for languages.}
		\label{fig:legend_langs}
	\end{subfigure}
	\qquad
	\begin{subfigure}{4cm}
	    \centering
		\includegraphics[width=4cm]{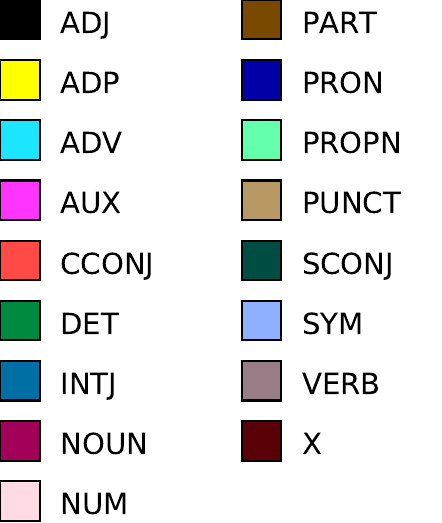}
		\caption{Colour legend for UDPOS labels.}
		\label{fig:legend_labels_udpos}
	\end{subfigure}
	\qquad
	\begin{subfigure}{4cm}
	    \centering
		\includegraphics[width=4cm]{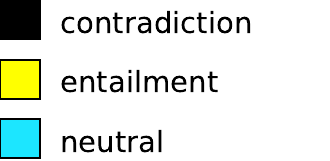}
		\caption{Colour legend for XNLI labels.}
		\label{fig:legend_labels_xnli}
	\end{subfigure}
 	\caption{Legends for t-SNE visualisations.}
	\label{fig:legend}
\end{figure*}

\input{fig_finetune_udpos}

\input{fig_finetune_xnli}

On the other hand, the opposite is observed for languages.
Here, mBERT starts off with fairly clear-cut language clusters which then become mixed (compare Figures \ref{fig:finetuning_udpos_frozen_label_lang} vs \ref{fig:finetuning_udpos_finetuned_label_lang}; and \ref{fig:finetuning_xnli_frozen_label_lang} vs \ref{fig:finetuning_xnli_finetuned_label_lang}).
This is less evident in the Wikipedia data set.
In general, we observe a loss of language-specific representational capacity following fine-tuning on both tasks, in line with the drop in LID performance in Table \ref{tab:finetuning_results}.

In the case of the Wikipedia plots for XNLI (Figures \ref{fig:finetuning_xnli_frozen_lang} and \ref{fig:finetuning_xnli_finetuned_lang}), there seems to be negligible clustering present (even according to V-measure) but a very high LID performance.
Given that the language classifier is single layer deep, this indicates that, although the representations are not clustered, they are still linearly separable (before being compressed by t-SNE).

With the exception of LID on Wikipedia after XNLI fine-tuning, there is a drop in LID performance, which we take as evidence that as mBERT is fine-tuned on a task, its representations become more language invariant.
Put differently, fine-tuning requires mBERT's finite representational capacity to be dedicated to the task requirements, at the expense of accurately distinguishing between languages.
In this sense, language-specific and language-neutral representations are competing with each other in the context of fine-tuning.

However, the results also show interesting differences between tasks: fine-tuning results in a steeper increase in F1 score for XNLI compared to UDPOS, suggesting that cross-lingual transfer on semantic tasks may benefit more if language specificity decreases, compared to tasks involving morphosyntax.
In a related vein, \citet{lauscher-etal-2020-zero} 
show that POS tagging and dependency parsing are impacted by structural language similarity.

\subsection{The impact of increasing language confusion}

In this section, we address whether, by using explicit means to enforce language confusion in mBERT representations, we can observe better performance on the target tasks in a cross-lingual zero-shot transfer setting.
Unlike the fine-tuning experiments, the language classifier in this case is not trained in isolation, but is trained together with the rest of the system in order to allow the mBERT model to learn to confuse the language classifier.
The language classifier is then retrained from scratch in isolation in order to be able to measure the amount of language sensitive information in the mBERT representations.

Table~\ref{tab:langunlearning_results} gives the macro F1 scores of the classifier predictions.
Compared to the simple fine-tuning setup in Table \ref{tab:finetuning_results}, it is clear that both gradient reversal and entropy maximisation have a negative impact on target task performance.

t-SNE plots for the mBERT representations after language unlearning are shown in the Appendix, but we show the V-measure values in Table~\ref{tab:langunlearning_vmeasure_results}.
In general, gradient reversal and entropy maximisation yield a comparable degree of clustering to that observed above after fine-tuning.
The exception is XNLI, where the V-measure is higher and suggests more well-defined clusters compared to the fine-tuned case.

\begin{table}[h]
	\centering
	\small
	\begin{tabular}{l|cc|cc} 
		\hline
		& \multicolumn{2}{c|}{\bf UDPOS} & \multicolumn{2}{c}{\bf XNLI} \\
		\hline
		& Grad. & Ent. & Grad. & Ent. \\
		\hline
		{\bf Target labels} &  22.8 & 30.1 & 1.1 & 12.2 \\
		\hline
		{\bf Language (Target)} & 12.5 & 12.1 & 18.8 & 20.0 \\
		{\bf Language (Wiki)} & 7.9 & 9.6 & 12.2 & 14.7 \\
		\hline
	\end{tabular}
	\caption{
		V-measures (\%) of full vector test set samples after clustering for target labels (UDPOS and XNLI) and languages after training using gradient reversal (Grad.) and entropy maximisation (Ent.).
		Note that `Language (Target)' refers to languages in the target data set.
	}
	\label{tab:langunlearning_vmeasure_results}
\end{table}

These results suggest that fine-tuning of multilingual models such as mBERT involves an interplay between language-specific and language-neutral representations.
They also indicate that the task matters: classifying tokens with morphosyntactic information results in greater loss of language specificity and in the ability to identify languages than classifying entire texts.
Furthermore, strategies to enforce language confusion in representations result in a deterioration of performance, interfering with the ability of the model to balance the two sources of information as a function of the task it is being fine-tuned on.

\section{Related work}

Several studies have shown that mBERT performs well in zero-shot transfer settings in low-level structured prediction tasks \cite{pires-etal-2019-multilingual} and in comparison to models that use explicit multilingual information for a variety of tasks and languages \cite{wu-dredze-2019-beto}.
Another focus of research has been the correlation between cross-lingual performance and shared vocabulary (e.g. as measured by wordpiece overlap).
\citet{Karthikeyan2020Cross-Lingual} find no such correlation, suggesting that mBERT's success in zero-shot transfer must be due to cross-lingual mappings at a deeper linguistic level.
However, it is not clear how this finding should be interpreted in light of the further finding that mBERT performs well (ca. 96\%) at language identification on the WiLI data set\footnote{
    \url{https://zenodo.org/record/841984}
}.

Other work has taken this further by focusing on the hypothesis that mBERT encodings contain both a language-specific and a language-neutral component \cite{libovicky-etal-2020-language}.
\citet{gonen-etal-2020-greek} set out to disentangle both components and find that in `language identity subspace', t-SNE projections show large improvement in clustering with respect to language.
In language-neutral space, semantic representations are largely intact.
\citet{dufter-schutze-2020-identifying} show that smaller models result in better cross-lingual zero-shot learning.
This could be due to there being less representational space allocated to language-specific information.

Some previous work has also shed light on the impact of training strategies on cross-lingual zero-shot learning.
\citet{Tan2021} show that training on code-switched data improves cross-lingual transfer, while \citet{Phang2020} show that intermediate task training, prior to fine-tuning proper, also results in better transfer with XLM-R \cite{conneau-etal-2020-unsupervised}.
This is related to the present finding that mBERT representations become less language-specific after fine-tuning.

In this paper, we also experiment with techniques that specifically enforce language confusion, using entropy maximisation or gradient reversal.
The latter has been used successfully for domain adaptation \cite{Ganin2015}, while \citet{libovicky-etal-2020-language} use it for additional pre-training.

\section{Conclusions}

This paper explored the interplay between language-specific and language-neutral representations in the multilingual transformer model mBERT.
Our results show that fine-tuning on specific tasks has a differential impact on how much language-specific information is retained.
On the other hand, gradient reversal and iterative entropy maximisation interfere with fine-tuning and do not improve task performance.

Fine-tuning for tasks requiring identification of morphosyntactic properties, such as POS tagging, can result in greater loss of language-specific information, compared to deeper semantically-oriented tasks such as NLI\footnote{
    It could be argued that the model's loss of language-specific information is not due to the nature of fine-tuning, but due to the fact that fine-tuning is carried out using monolingual data (English).
    This could, in principle, imply that the model is not being given any indication that language-specific information should be retained.
    However, we find this an unlikely explanation since it does not account for the substantial difference in LID performance after fine-tuning on {\em different} tasks.
}.

However, there are also differences in the input data of these two tasks which could have impacted the results reported here.
The data sets differ in their homogeneity (unlike XNLI, UDPOS is constructed from multiple sources) and size.
The tasks also differ in their granularity, in that NLI is text-level, while POS is token-level.
Furthermore, NLI is known to be susceptible to `shortcut learning', whereby models rely on recurrent biases in training to solve the task \cite{DAmour2020} whereas shortcuts in part of speech tagging are less known.
These considerations are worthy of further investigation.

We believe these results contribute towards greater understanding of fine-tuning on multilingual model representations.
This results in a reorganisation of the representation space, suggesting that language-specific and language-independent subspaces are dependent on task.

\section{Ethical considerations}

The experiments reported here rely on previously available data sets and/or data extracted from publicly available sources.
To ensure reproducibility, we will release all code, including scripts to regenerate the Wikipedia data set for language identification.

\section*{Acknowledgements}

This research was supported by a Malta Enterprise Research and Development grant.
We thank our collaborators, CityFalcon Ltd.\footnote{\url{https://www.cityfalcon.com/}}
Comments and questions by four anonymous reviewers are also gratefully acknowledged.

\bibliography{bibliography}
\bibliographystyle{acl_natbib}

\clearpage
\section*{Appendix}
\input{appendix}

\end{document}

%% file: fig_finetune_udpos.tex
\begin{figure*}
    \centering
	\begin{subfigure}{0.45\textwidth}
	    \centering
		\includegraphics[width=5cm]{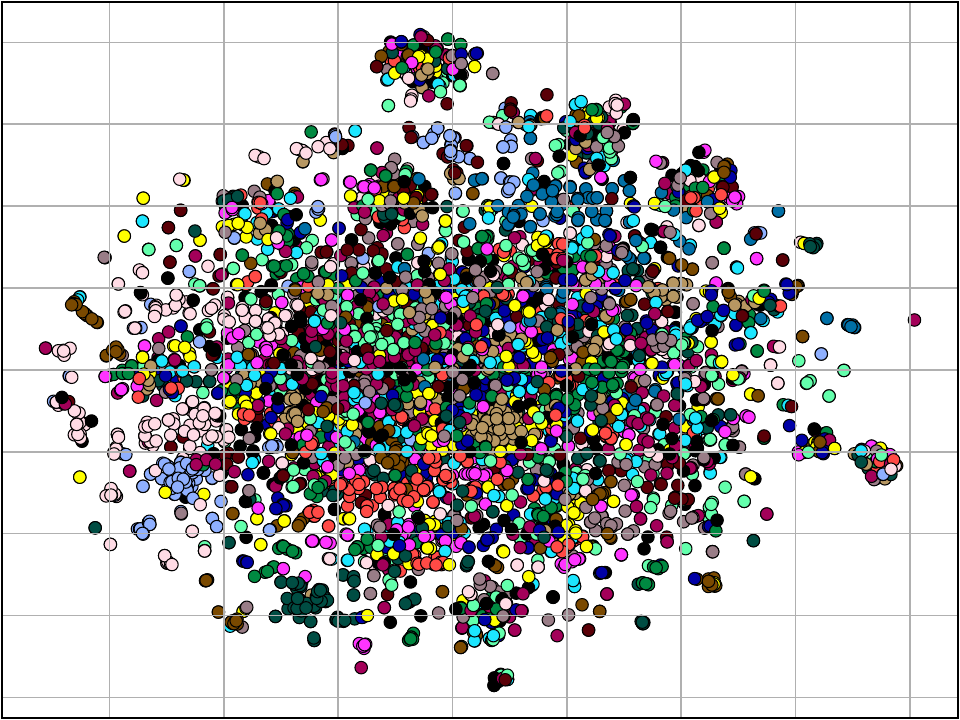}
		\caption{
			UDPOS labels with initial mBERT (macro F1: 51.2\%, V-measure: 13.0\%).
		}
		\label{fig:finetuning_udpos_frozen_label_label}
	\end{subfigure}
	\qquad
	\begin{subfigure}{0.45\textwidth}
	    \centering
		\includegraphics[width=5cm]{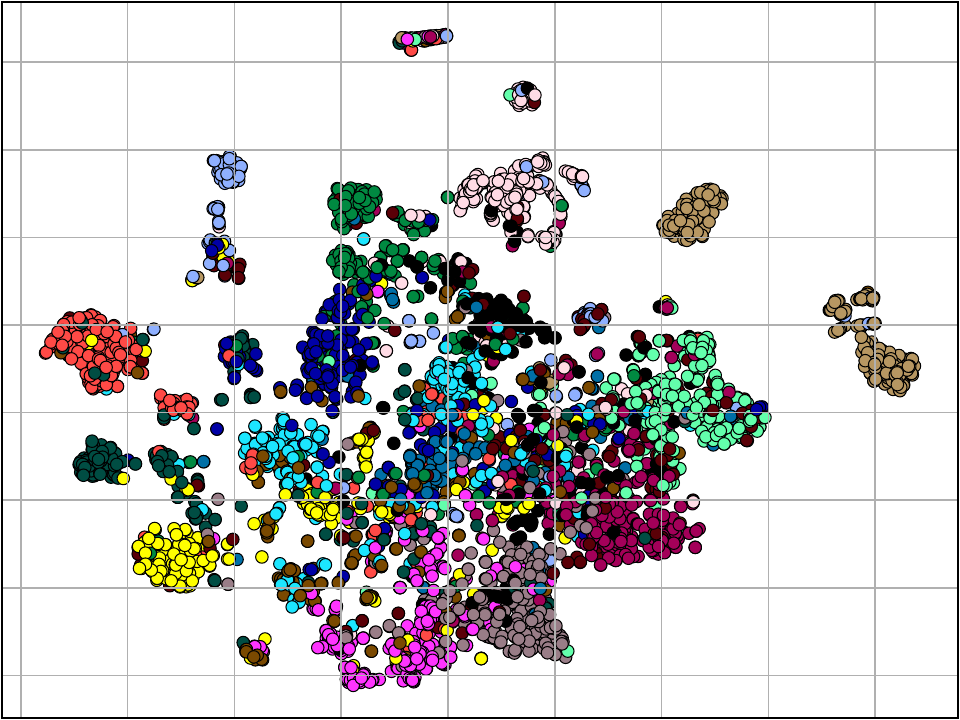}
		\caption{
			UDPOS labels with fine-tuned mBERT (macro F1: 59.6\%, V-measure: 47.5\%).
		}
		\label{fig:finetuning_udpos_finetuned_label_label}
	\end{subfigure}

	\begin{subfigure}{0.45\textwidth}
	    \centering
		\includegraphics[width=5cm]{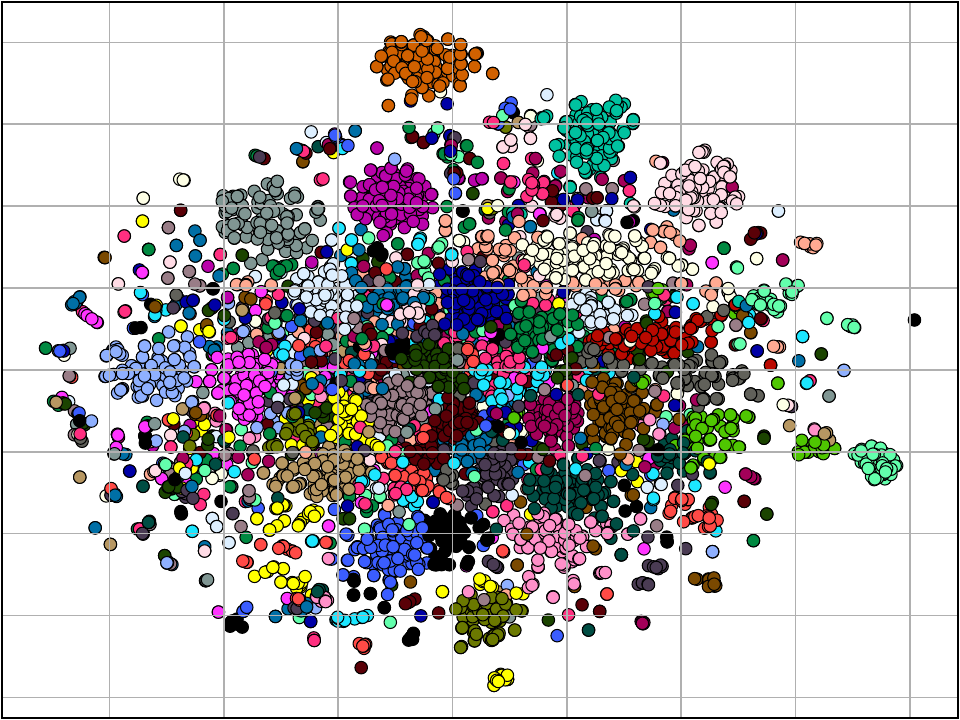}
		\caption{
			UDPOS languages with initial mBERT (macro F1: 78.3\%, V-measure: 49.2\%).
		}
		\label{fig:finetuning_udpos_frozen_label_lang}
	\end{subfigure}
	\qquad
	\begin{subfigure}{0.45\textwidth}
	    \centering
		\includegraphics[width=5cm]{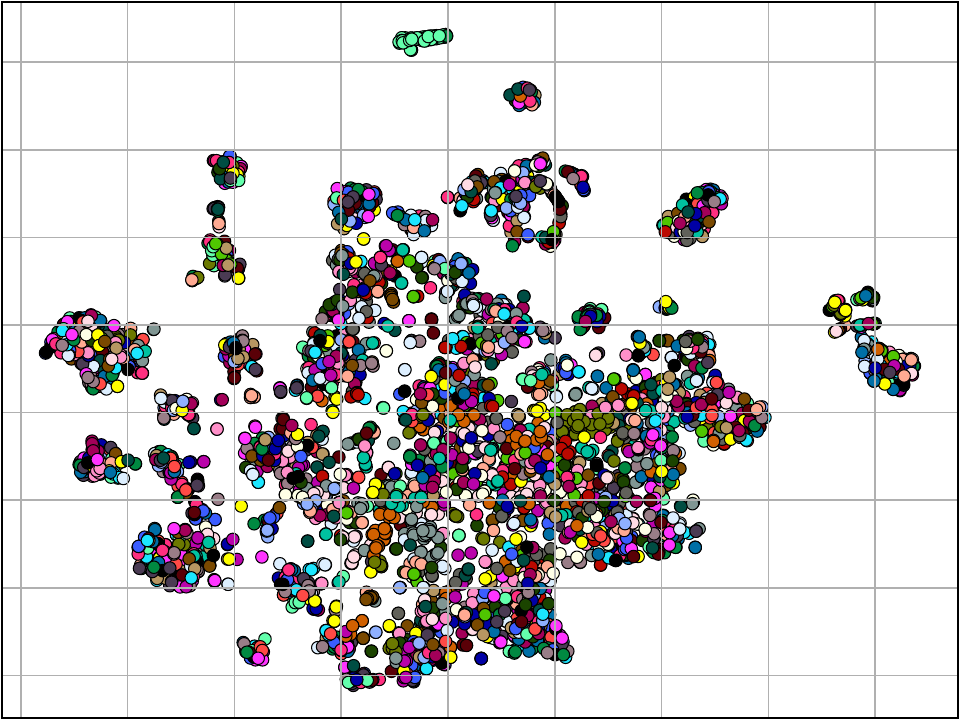}
		\caption{
			UDPOS languages with fine-tuned mBERT (macro F1: 0.3\%, V-measure: 10.4\%).
		}
		\label{fig:finetuning_udpos_finetuned_label_lang}
	\end{subfigure}

	\begin{subfigure}{0.45\textwidth}
	    \centering
		\includegraphics[width=5cm]{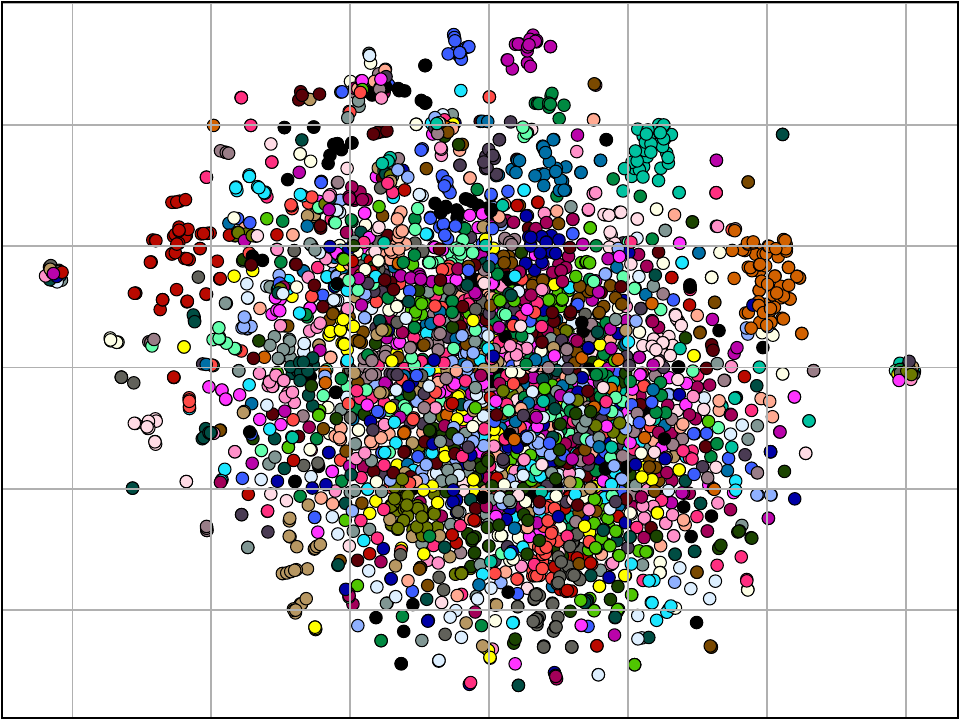}
		\caption{
			Wikipedia languages with initial mBERT (macro F1: 59.3\%, V-measure: 22.3\%).
		}
		\label{fig:finetuning_udpos_frozen_lang}
	\end{subfigure}
	\qquad
	\begin{subfigure}{0.45\textwidth}
	    \centering
		\includegraphics[width=5cm]{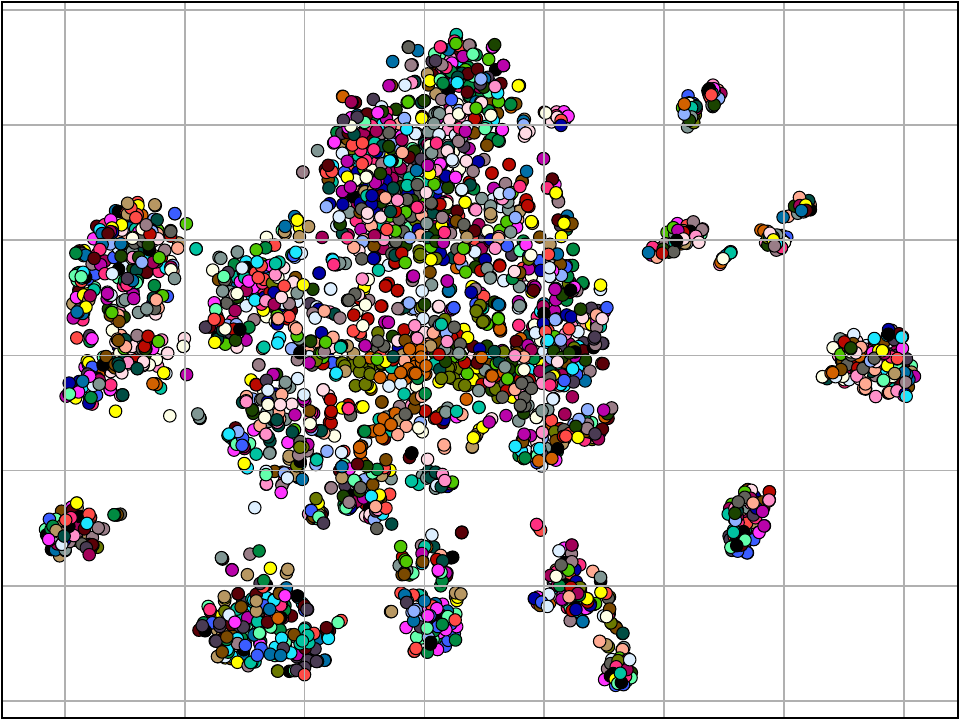}
		\caption{
			Wikipedia languages with fine-tuned mBERT (macro F1: 0.5\%, V-measure: 11.6\%).
		}
		\label{fig:finetuning_udpos_finetuned_lang}
	\end{subfigure}

	\caption{
	    2D t-SNE projections before and after fine-tuning, with UDPOS as target task.
	    Macro F1 scores for label/language classification of mBERT embeddings are also given.
	    Points represent tokens.
	    See Figure~\ref{fig:legend_langs} (languages) and Figure~\ref{fig:legend_labels_udpos} (labels) for colour legends.
    }
	\label{fig:finetuning_udpos_plots}
\end{figure*}

%% file: fig_finetune_xnli.tex
\begin{figure*}[t]
    \centering
	\begin{subfigure}{0.45\textwidth}
	    \centering
		\includegraphics[width=5cm]{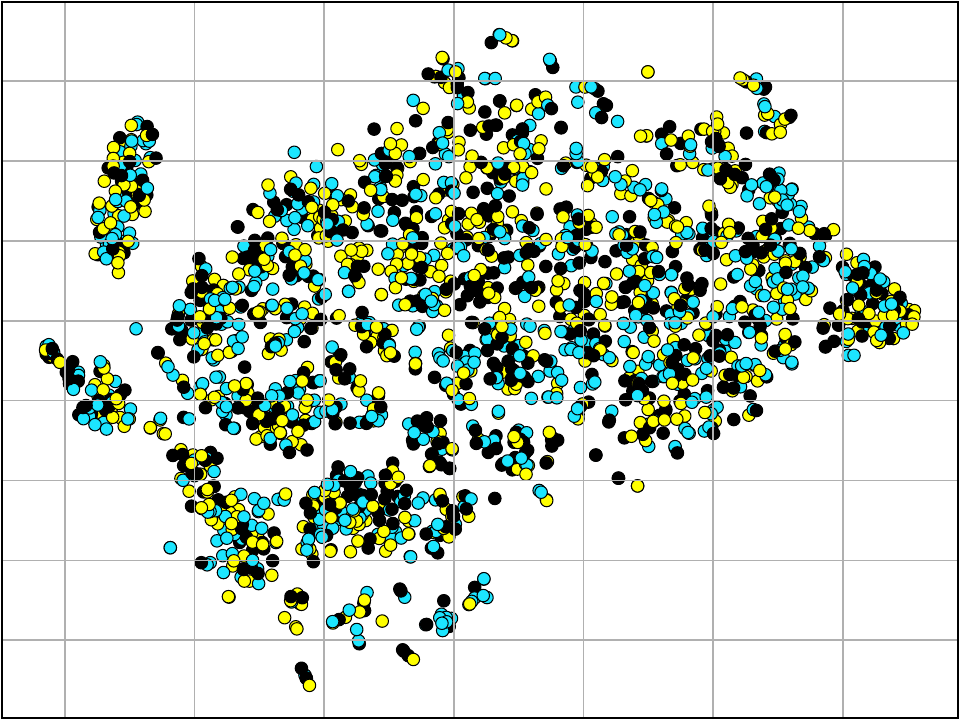}
		\caption{
			XNLI labels with initial mBERT (macro F1: 29.7\%, V-measure: 0.1\%).
		}
		\label{fig:finetuning_xnli_frozen_label_label}
	\end{subfigure}
	\qquad
	\begin{subfigure}{0.45\textwidth}
	    \centering
		\includegraphics[width=5cm]{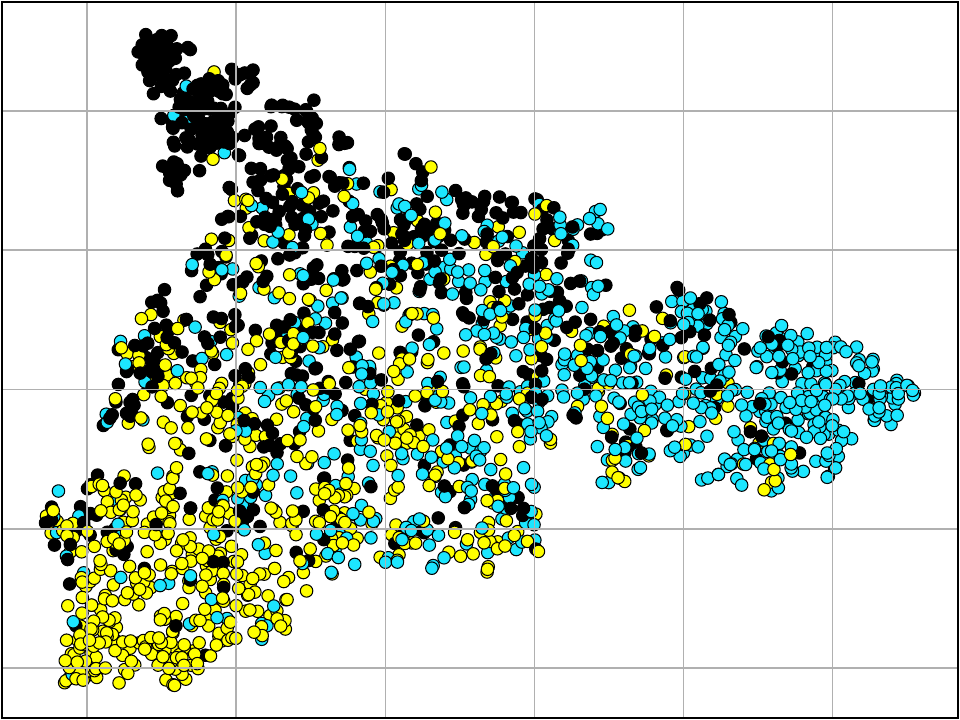}
		\caption{
			XNLI labels with fine-tuned mBERT (macro F1: 66.3\%, V-measure: 20.3\%).
		}
		\label{fig:finetuning_xnli_finetuned_label_label}
	\end{subfigure}
	
	\begin{subfigure}{0.45\textwidth}
	    \centering
		\includegraphics[width=5cm]{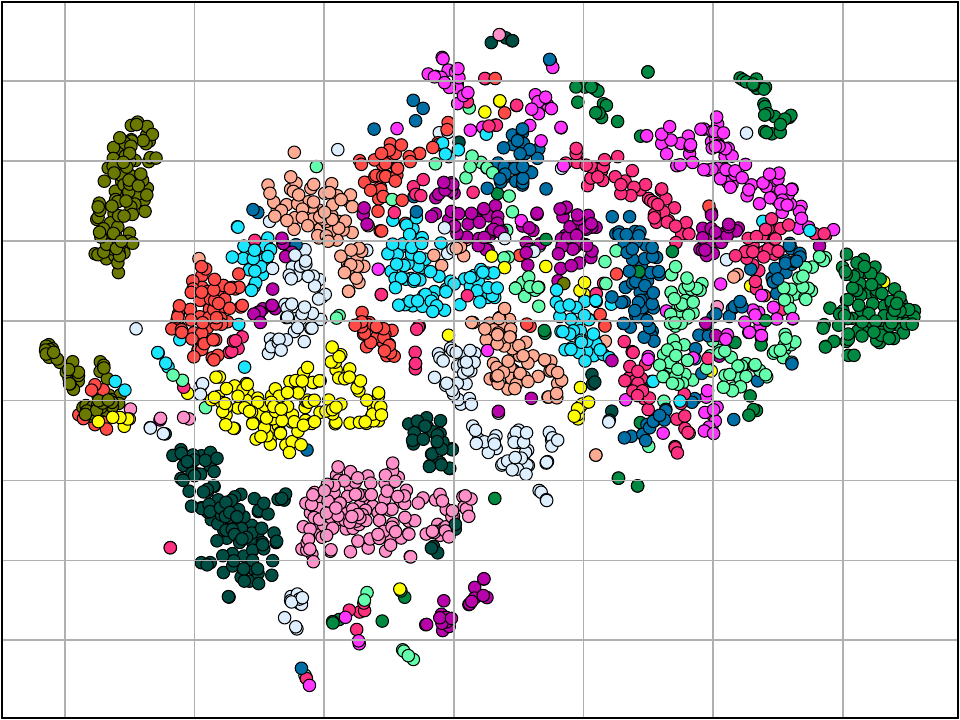}
		\caption{
			XNLI languages with initial mBERT (macro F1: 49.8\%, V-measure: 35.1\%).
		}
		\label{fig:finetuning_xnli_frozen_label_lang}
	\end{subfigure}
	\qquad
	\begin{subfigure}{0.45\textwidth}
	    \centering
		\includegraphics[width=5cm]{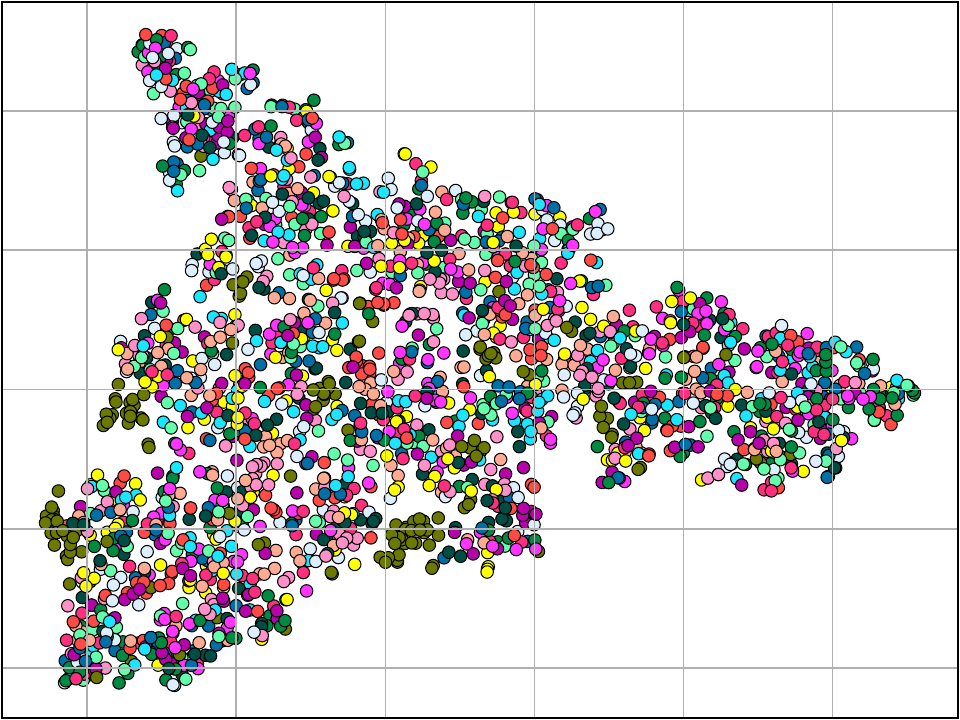}
		\caption{
			XNLI languages with fine-tuned mBERT (macro F1: 39.2\%, V-measure: 6.8\%).
		}
		\label{fig:finetuning_xnli_finetuned_label_lang}
	\end{subfigure}
	
	\begin{subfigure}{0.45\textwidth}
	    \centering
		\includegraphics[width=5cm]{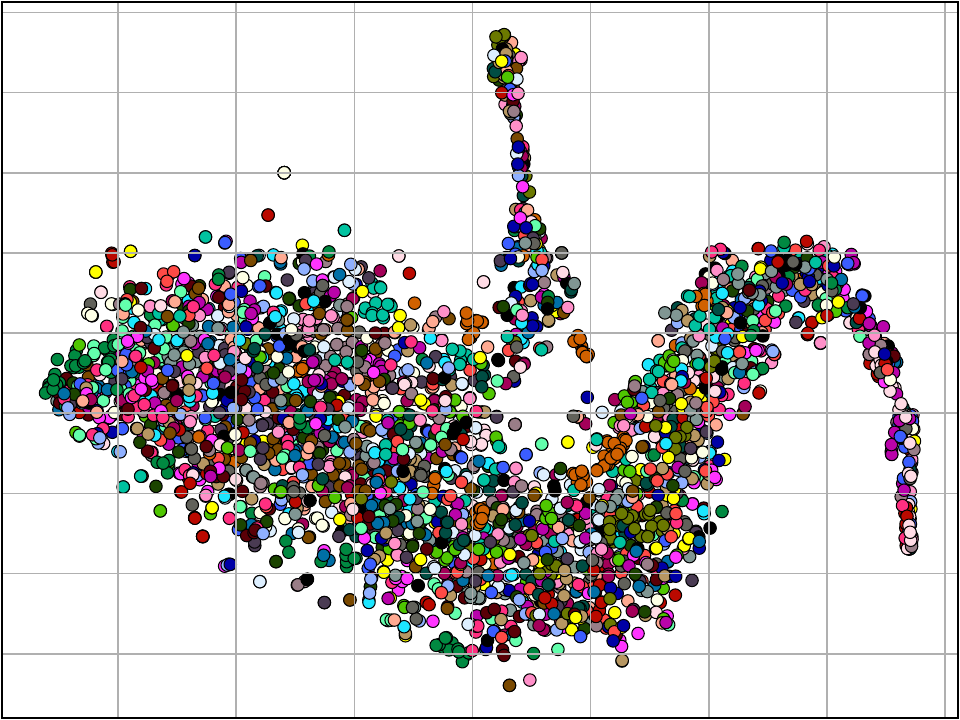}
		\caption{
			Wikipedia languages with initial mBERT (macro F1: 97.0\%, V-measure: 11.7\%).
		}
		\label{fig:finetuning_xnli_frozen_lang}
	\end{subfigure}
	\qquad
	\begin{subfigure}{0.45\textwidth}
	    \centering
		\includegraphics[width=5cm]{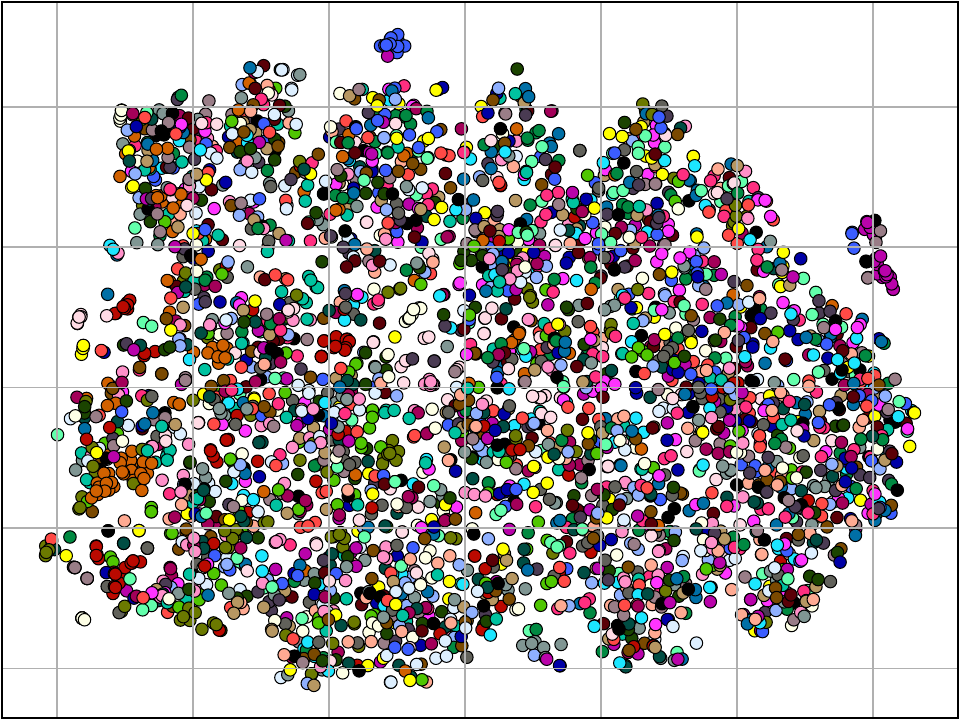}
		\caption{
			Wikipedia languages with fine-tuned mBERT (macro F1: 97.2\%, V-measure: 8.9\%).
		}
		\label{fig:finetuning_xnli_finetuned_lang}
	\end{subfigure}
	
	\caption{
		2D t-SNE projections before and after fine-tuning, with XNLI as target task.
		Macro F1 scores for label/language classification of mBERT embeddings are also given.
		Points represent pairs of sentences.
		See Figure~\ref{fig:legend_langs} (languages) and Figure~\ref{fig:legend_labels_xnli} (labels) for colour legends.
	}
	\label{fig:finetuning_xnli_plots}
\end{figure*}

%% file: appendix.tex
\appendix

\section{Data set details}
\label{sec:datastats}

When splitting data sets, a stratified split was used to preserve the proportions of different languages between splits.

\paragraph{UDPOS} After filtering data so that only the 33 selected languages are used, we split the initial training set into a smaller training set (90\%) and a validation set (10\%).
Non-English data was removed from the training set and validation set.
The training set contains 18\,906 sentences, the validation set contains 2\,100 sentences, development set contains 66\,062, and the test set contains 95\,065 sentences.

\paragraph{XNLI} We split the training set (which contains only English text) into a smaller training set (90\%) and a validation set (10\%).
The training set contains 351\,340 pairs of sentences, the validation set contains 39\,037 pairs, development set contains 33\,484, and the test set contains 67\,459 pairs.

\paragraph{Wikipedia} After randomly sampling 5\,000 paragraphs per language from the 33 selected languages, we split the paragraphs into train/val/dev/test splits using 70/10/10/10\% splits.
The training set contains 115\,500 paragraphs, the validation set contains 16\,500 paragraphs, development set contains 16\,500, and the test set contains 16\,500 paragraphs.
Text was extracted from the {\tt 20200420} Wikimedia dump using {\tt wikiextractor}.\footnote{
    \url{https://github.com/attardi/wikiextractor}.
}

\section{t-SNE/V-measure data sample}
\label{sec:data_sample}

For t-SNE plots and V-measure computation, we use a random sample of data points.
The samples were chosen such that a maximum number of points is shown from each label-language pair in the target data set, or from each language in the Wikipedia data set.
Since different data sets have different numbers of label-language pairs, a different maximum is chosen for each data set in order to keep them similarly sized in total.

For UDPOS, which has the largest number of language-label pairs, only a maximum of 10 points are shown from each pair.
This results in 5\,013 points in total.

XNLI has three different labels, so a maximum of 50 points is shown from each pair.
This results in 2\,100 points in total.

Wikipedia has no labels and a maximum of 100 points is shown from each language.
This results in 3\,300 points in total.

\section{Model training details}
\label{sec:model_training}

All experiments are conducted on a 64GB RAM server with an Intel(R) Xeon W-2123 8-core CPU (3.60GHz) and a GeForce Titan RTX2080 Ti GPU.

Experiments are conducted using PyTorch and the implementation of Multilingual BERT in the {\tt transformers} Python library.\footnote{
    \url{https://huggingface.co/bert-base-multilingual-cased}
}
We use the Adam optimiser with a different learning rate for the mBERT model and for the classification layers, although both classification layers use the same learning rate.

The classification layers are initialised randomly using a normal distribution with mean zero.
The mBERT encodings are passed through a dropout layer with a dropout rate of 0.1 before being passed to the classification layers.

We train the model for 5 epochs.
A validation set is used to measure the model's performance after each epoch; we reserve the model at the best epoch out of the five.

When optimising the label classifier, the macro F1 score performance on the English data of the target task validation set is measured (only English is considered in order to be faithful to a cross-lingual zero-shot learning setting).

On the other hand, when optimising the language classifier, the macro F1 score performance on all of the Wikipedia validation set is measured (language labels are assumed to be available during zero-shot learning).

In the gradient reversal experiments, only the target task validation set is used for determining which epoch gave the best model.
Two minibatches of the same size from the labelled data and the language data are used for each parameter update.

\section{Hyperparameter tuning}
\label{sec:hyperparameter_tuning}

Hyperparameter tuning is conducted via random search over a fixed set of values per hyperparameter.
Due to time constraints, only 20 random sets of hyperparameters are sampled for each experiment.
The development set is used to evaluate hyperparameters during hyperparameter tuning and only the label classifier is evaluated.
Hyperparameters that best fit the label classifier are also used in the language classifier.

The hyperparameter values used in the random search are given in Table~\ref{tab:hyperparameter_ranges}.
Selected hyperparameter values for each model are given in Table~\ref{tab:hyperparameters}.

\begin{table}
    \centering
    \begin{tabular}{c|c}
        Hyperparameter & Values \\
        \hline
        Init. standard deviation & 1e-1, 1e-2, 1e-3 \\
        Minibatch size & 64, 32, 16 \\
        Classifier learning rate & 1e-1, 1e-2, 1e-3, 1e-4 \\
        mBERT learning rate & 1e-3, 1e-4, 1e-5, 1e-6 \\
        Gradient reversal lambda & 0.1, 0.3, 0.5, 0.7 \\
        Entropy max. weighting & 0.1, 0.3, 0.5, 0.7 \\
    \end{tabular}
    \caption{Sets of hyperparameter values sampled from during hyperparameter tuning.}
    \label{tab:hyperparameter_ranges}
\end{table}

\begin{table*}
    \centering
    \begin{tabular}{c|p{11cm}}
        Experiment & Hyperparameters \\
        \hline
        UDPOS frozen & init. stddev.: 1e-01, minibatch size: 16, class. learning rate: 1e-03 \\
        UDPOS fine-tuned & init. stddev.: 1e-02, minibatch size: 64, mBERT learning rate: 1e-04, class. learning rate: 1e-01 \\
        UDPOS grad. rev. & init. stddev.: 1e-03, minibatch size: 32, mBERT learning rate: 1e-06, class. learning rate: 1e-03, grad. rev. $\lambda$: 0.1 \\
        UDPOS ent. max. & init. stddev.: 1e-02, minibatch size: 32, mBERT learning rate: 1e-06, class. learning rate: 1e-02, entropy max. $w$: 0.7 \\
        XNLI frozen & init. stddev.: 1e-02, minibatch size: 64, class. learning rate: 1e-02 \\
        XNLI fine-tuned & init. stddev.: 1e-03, minibatch size: 64, mBERT learning rate: 1e-05, class. learning rate: 1e-02 \\
        XNLI grad. rev. & init. stddev.: 1e-03, minibatch size: 32, mBERT learning rate: 1e-06, class. learning rate: 1e-03, grad. rev. $\lambda$: 0.1 \\
        XNLI ent. max. & init. stddev.: 1e-01, minibatch size: 32, mBERT learning rate: 1e-06, class. learning rate: 1e-04, entropy max. $w$: 0.1 \\
    \end{tabular}
    \caption{Selected hyperparameter values used in training.}
    \label{tab:hyperparameters}
\end{table*}

\section{Results on XNLI after language unlearning}
\label{sec:langunlearning_xnli}

In Figure~\ref{fig:langunlearning_udpos_plots} and Figure~\ref{fig:langunlearning_xnli_plots} we reproduce the visualisations after gradient reversal and entropy maximisation, when the model is fine-tuned on UDPOS and XNLI respectively.

\input{fig_unlearn_udpos}

\input{fig_unlearn_xnli}

%% file: fig_unlearn_udpos.tex
\begin{figure*}
	\centering
	\begin{subfigure}{0.45\textwidth}
    	\centering
		\includegraphics[width=5cm]{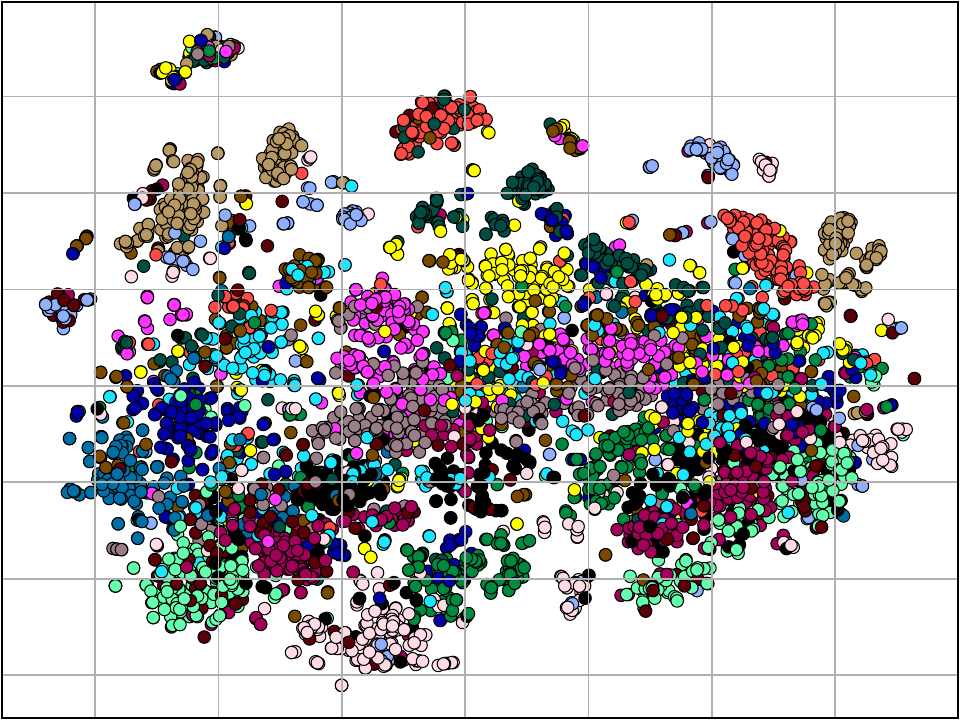}
		\caption{
			UDPOS labels with gradient reversal (macro F1: 53.5\%, V-measure: 22.8\%).
		}
		\label{fig:langunlearning_udpos_frozen_label_label}
	\end{subfigure}
	\qquad
	\begin{subfigure}{0.45\textwidth}
	    \centering
		\includegraphics[width=5cm]{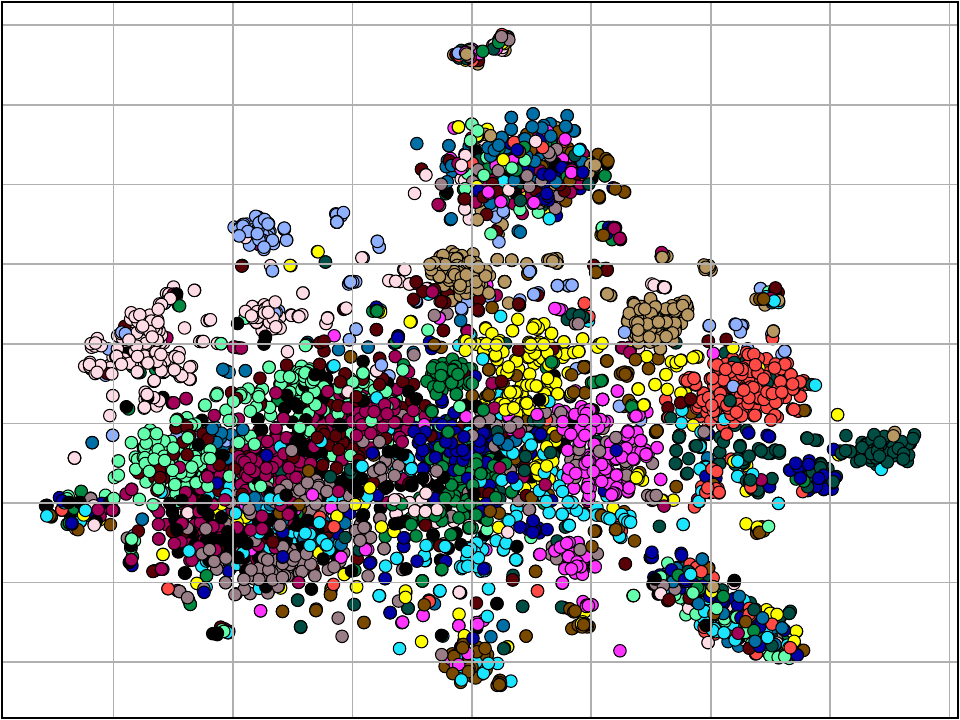}
		\caption{
			UDPOS labels with entropy maximisation (macro F1: 56.8\%, V-measure: 30.1\%).
		}
		\label{fig:langunlearning_udpos_finetuned_label_label}
	\end{subfigure}
	
	\begin{subfigure}{0.45\textwidth}
	    \centering
		\includegraphics[width=5cm]{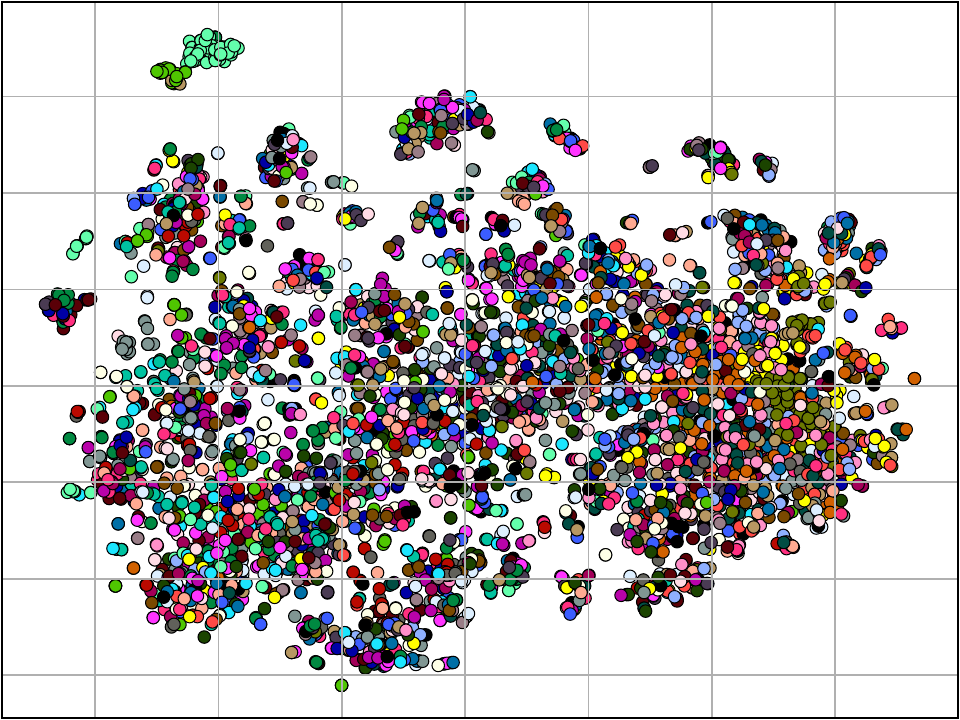}
		\caption{
			UDPOS language clusters after gradient reversal (macro F1: 0.1\%, V-measure: 12.5\%).
		}
		\label{fig:langunlearning_udpos_frozen_label_lang}
	\end{subfigure}
	\qquad
	\begin{subfigure}{0.45\textwidth}
	    \centering
		\includegraphics[width=5cm]{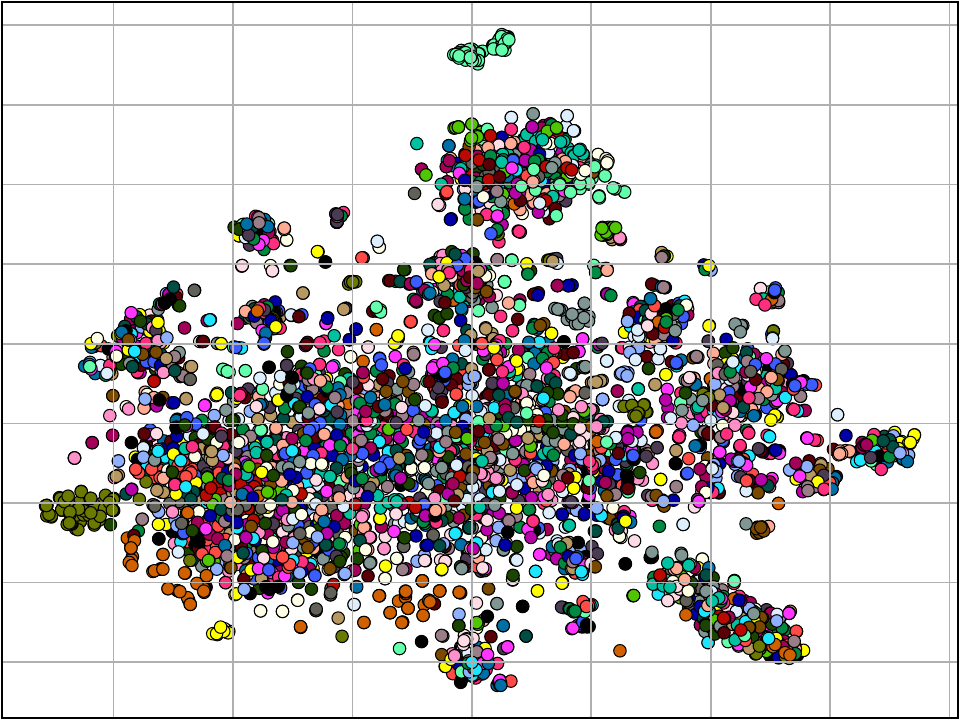}
		\caption{
			UDPOS language clusters after entropy maximisation learned mBERT (macro F1: 5.5\%, V-measure: 12.1\%).
		}
		\label{fig:langunlearning_udpos_finetuned_label_lang}
	\end{subfigure}
	
 	\begin{subfigure}{0.45\textwidth}
 	    \centering
 		\includegraphics[width=5cm]{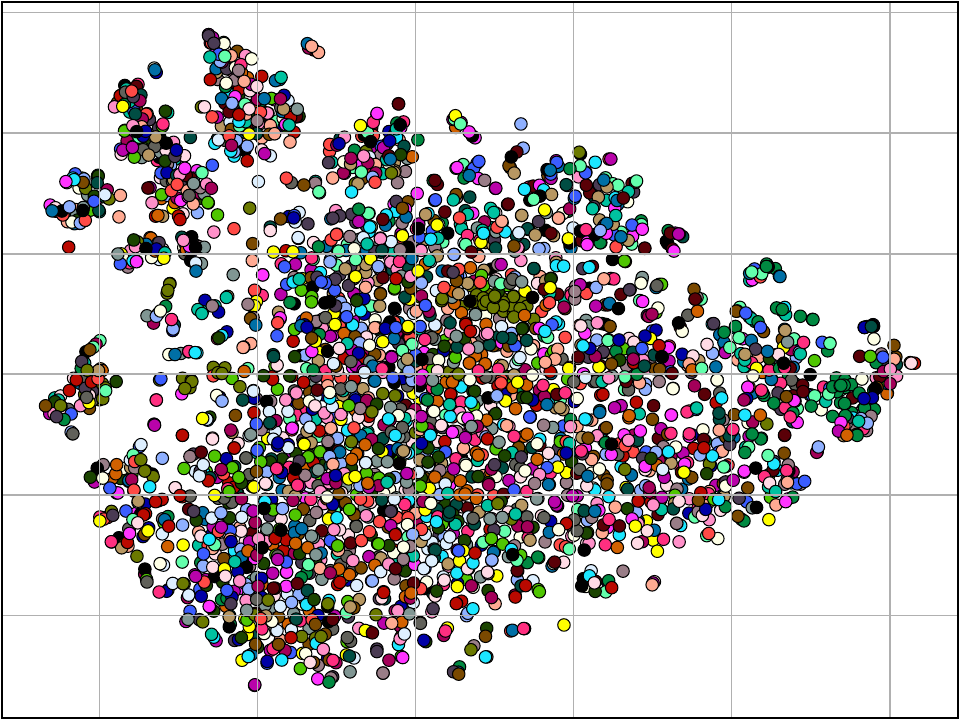}
 		\caption{
 			Wikipedia language clusters after gradient reversal (macro F1: 0.1\%, V-measure: 7.9\%).
 		}
 		\label{fig:langunlearning_udpos_frozen_lang}
 	\end{subfigure}
 	\qquad
 	\begin{subfigure}{0.45\textwidth}
 	    \centering
 		\includegraphics[width=5cm]{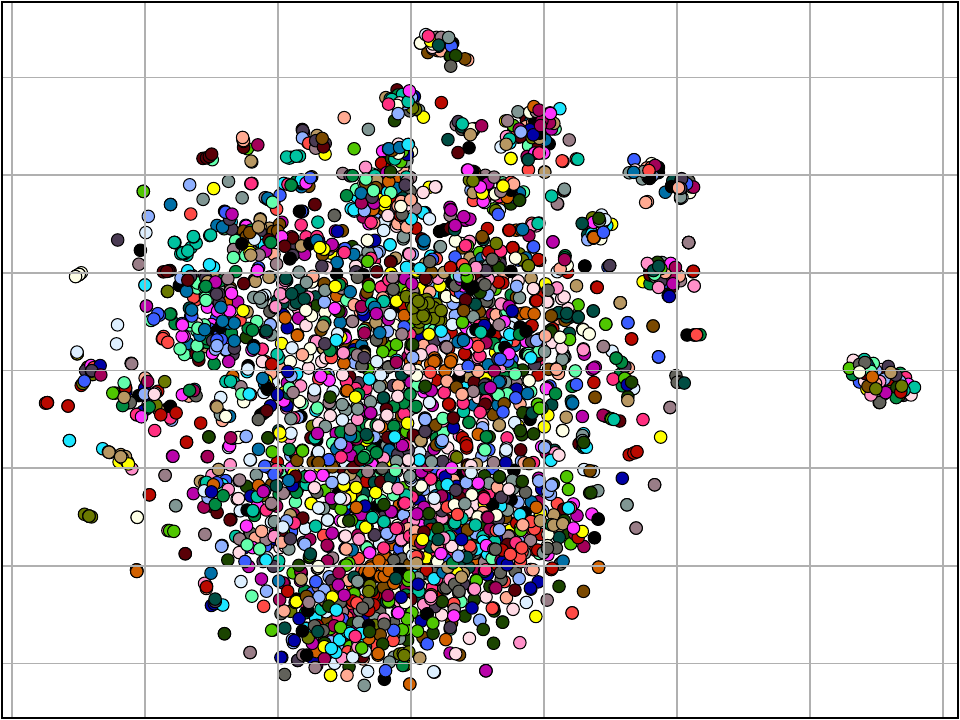}
 		\caption{
 			Wikipedia language clusters after entropy maximisation (macro F1: 3.1\%, V-measure: 9.6\%).
 		}
 		\label{fig:langunlearning_udpos_finetuned_lang}
 	\end{subfigure}
	
	\caption{
		Token based t-SNE plots of target task label and language clusters, after fine-tuning on UDPOS with gradient reversal or entropy maximisation.
		Each point is a randomly sampled token. 
		Macro F1 scores for classification of mBERT embeddings is also given.
	}
	\label{fig:langunlearning_udpos_plots}
\end{figure*}

%% file: fig_unlearn_xnli.tex
\begin{figure*}
 	\centering
	\begin{subfigure}{0.45\textwidth}
        \centering
		\includegraphics[width=5cm]{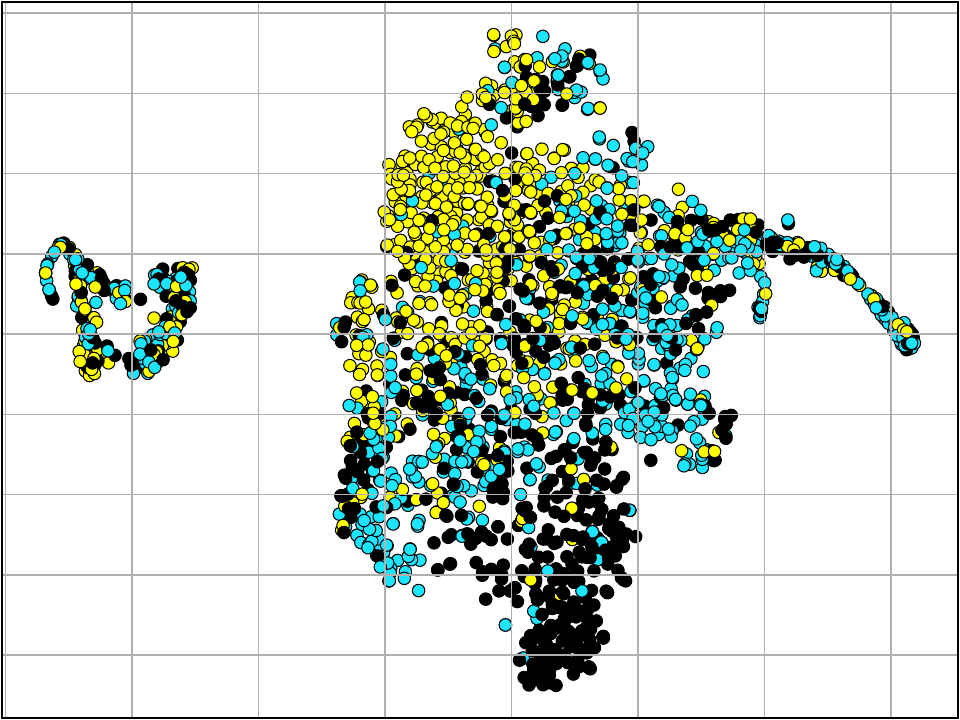}
		\caption{
			XNLI labels with gradient reversal (macro F1: 62.2\%, V-measure: 1.1\%).
		}
		\label{fig:langunlearning_xnli_frozen_label_label}
	\end{subfigure}
	\qquad
	\begin{subfigure}{0.45\textwidth}
	    \centering
		\includegraphics[width=5cm]{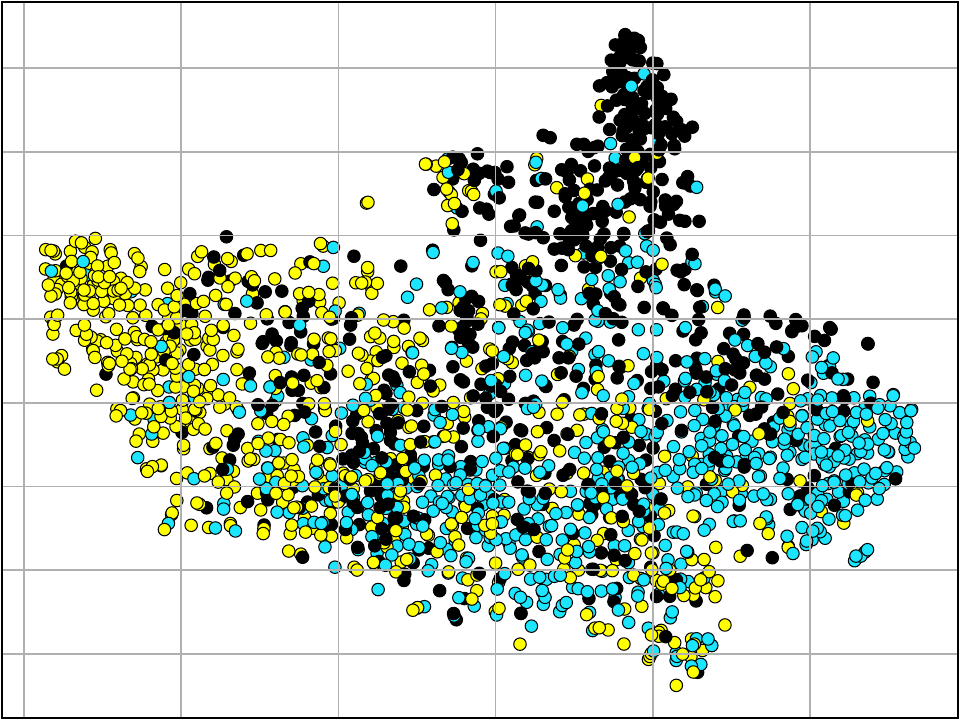}
		\caption{
			XNLI labels with entropy maximisation (macro F1: 62.1\%, V-measure: 12.2\%).
		}
		\label{fig:langunlearning_xnli_finetuned_label_label}
	\end{subfigure}
	
	\begin{subfigure}{0.45\textwidth}
	    \centering
		\includegraphics[width=5cm]{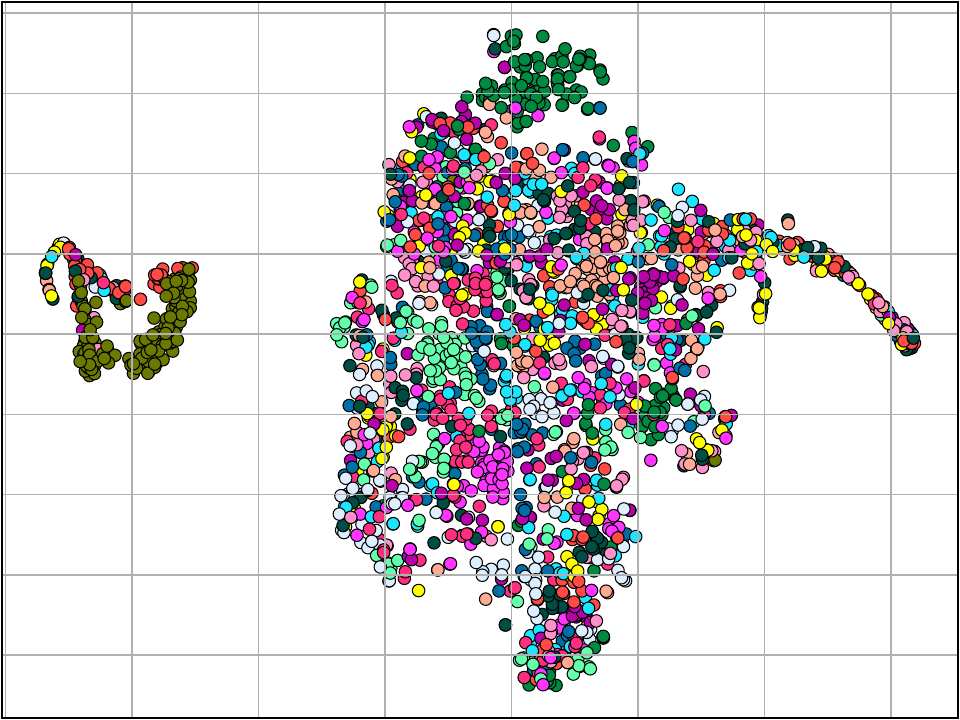}
		\caption{
			XNLI languages with gradient reversal (macro F1: 1.3\%, V-measure: 18.8\%).
		}
		\label{fig:langunlearning_xnli_frozen_label_lang}
	\end{subfigure}
	\qquad
	\begin{subfigure}{0.45\textwidth}
	    \centering
		\includegraphics[width=5cm]{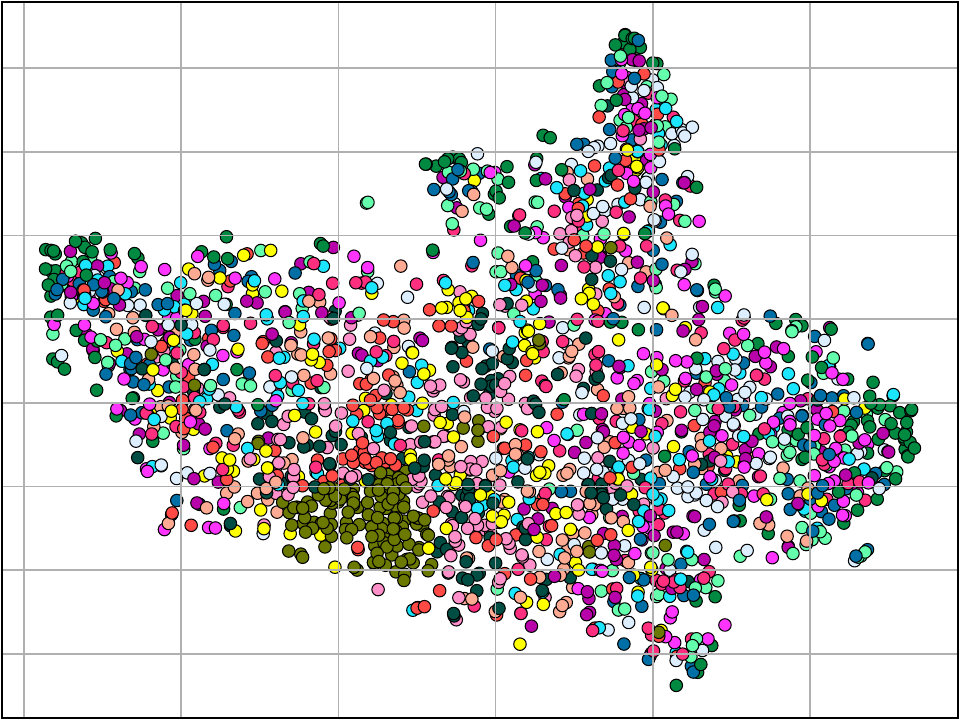}
		\caption{
			XNLI languages with entropy maximisation (macro F1: 3.4\%, V-measure: 20.0\%).
		}
		\label{fig:langunlearning_xnli_finetuned_label_lang}
	\end{subfigure}
	
 	\begin{subfigure}{0.45\textwidth}
 	    \centering
 		\includegraphics[width=5cm]{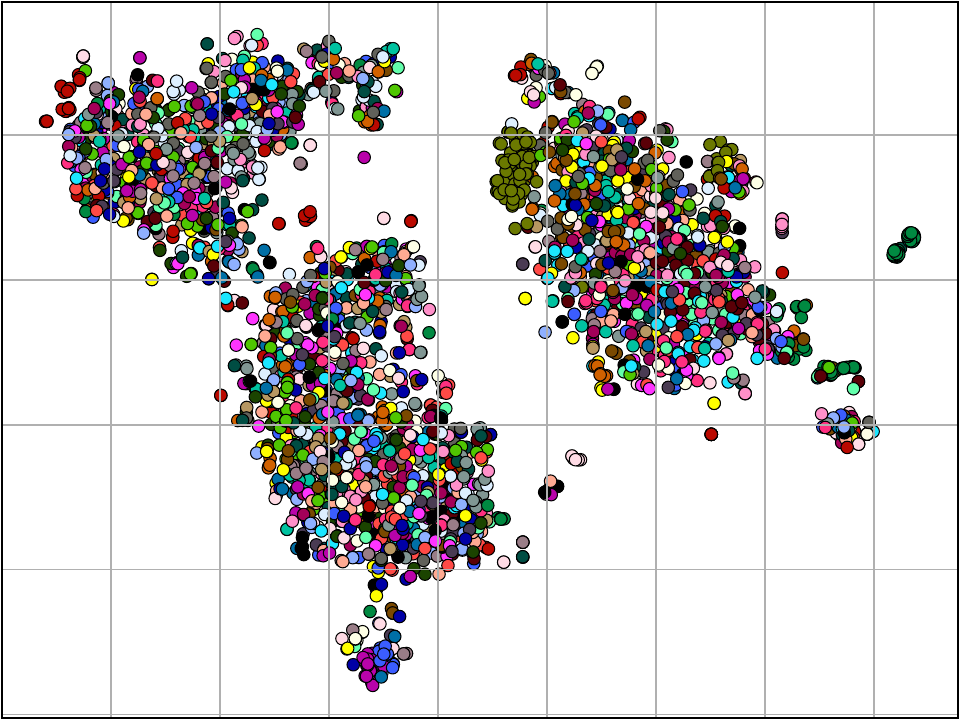}
 		\caption{
 			Wikipedia languages with gradient reversal learned mBERT (macro F1: 1.5\%, V-measure: 12.2\%).
 		}
 		\label{fig:langunlearning_xnli_frozen_lang}
 	\end{subfigure}
 	\qquad
 	\begin{subfigure}{0.45\textwidth}
 	    \centering
 		\includegraphics[width=5cm]{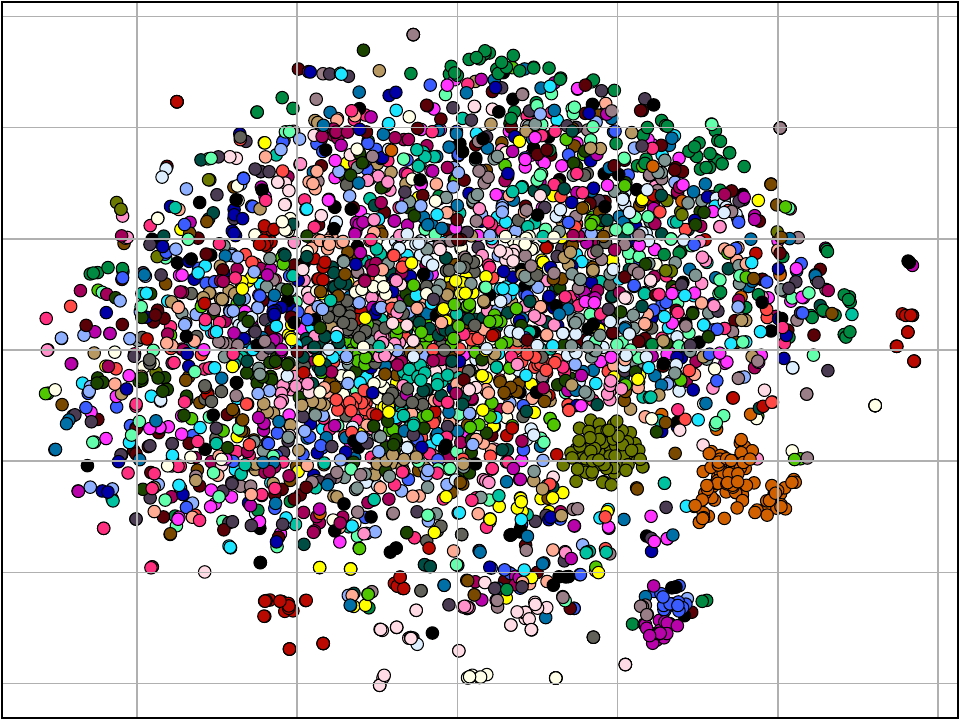}
 		\caption{
 			Wikipedia languages with entropy maximisation learned mBERT (macro F1: 54.3\%, V-measure: 14.7\%).
 		}
 		\label{fig:langunlearning_xnli_finetuned_lang}
 	\end{subfigure}
	
	\caption{
	    Text-based t-SNE plots of language clusters, after fine-tuning on XNLI with gradient reversal or entropy maximisation.
	    Macro F1 scores for classification of mBERT embeddings is also given.
	}
	\label{fig:langunlearning_xnli_plots}
\end{figure*}

%% file: ms.bbl
\begin{thebibliography}{27}
\expandafter\ifx\csname natexlab\endcsname\relax\def\natexlab#1{#1}\fi

\bibitem[{Artetxe et~al.(2019)Artetxe, Ruder, and Yogatama}]{Artetxe2019}
Mikel Artetxe, Sebastian Ruder, and Dani Yogatama. 2019.
\newblock \href {https://doi.org/10.18653/v1/2020.acl-main.421} {{On the
  cross-lingual transferability of monolingual representations}}.
\newblock In \emph{Proceedings ofthe 58th Annual Meeting ofthe Association for
  Computational Linguistics (ACL'20)}, pages 4623--4637. Association for
  Computational Linguistics.

\bibitem[{Bowman et~al.(2015)Bowman, Angeli, Potts, and Manning}]{Bowman2015}
Samuel~R Bowman, Gabor Angeli, Christopher Potts, and Christopher~D Manning.
  2015.
\newblock {A large annotated corpus for learning natural language inference}.
\newblock In \emph{Proceedings of the 2015 Conference on Empirical Methods in
  Natural Language Processing}, pages 632--642, Lisbon, Portugal. Association
  for Computational Linguistics.

\bibitem[{Choenni and Shutova(2020)}]{ChoenniWhat2020}
Rochelle Choenni and Ekaterina Shutova. 2020.
\newblock \href {http://arxiv.org/abs/2009.12862} {What does it mean to be
  language-agnostic? probing multilingual sentence encoders for typological
  properties}.
\newblock \emph{CoRR}, abs/2009.12862.

\bibitem[{Conneau et~al.(2020{\natexlab{a}})Conneau, Khandelwal, Goyal,
  Chaudhary, Wenzek, Guzman, Grave, Ott, Zettlemoyer, and
  Stoyanov}]{Conneau2020}
Alexis Conneau, Kartikay Khandelwal, Naman Goyal, Vishrav Chaudhary, Guillaume
  Wenzek, Francisco Guzman, Edouard Grave, Myle Ott, Luke Zettlemoyer, and
  Veselin Stoyanov. 2020{\natexlab{a}}.
\newblock {Unsupervised Cross-lingual Representation Learning at Scale}.
\newblock In \emph{Proceedings ofthe 58th Annual Meeting ofthe Association for
  Computational Linguistics (ACL'20)}, pages 8440--8451.

\bibitem[{Conneau et~al.(2020{\natexlab{b}})Conneau, Khandelwal, Goyal,
  Chaudhary, Wenzek, Guzm{\'a}n, Grave, Ott, Zettlemoyer, and
  Stoyanov}]{conneau-etal-2020-unsupervised}
Alexis Conneau, Kartikay Khandelwal, Naman Goyal, Vishrav Chaudhary, Guillaume
  Wenzek, Francisco Guzm{\'a}n, Edouard Grave, Myle Ott, Luke Zettlemoyer, and
  Veselin Stoyanov. 2020{\natexlab{b}}.
\newblock \href {https://doi.org/10.18653/v1/2020.acl-main.747} {Unsupervised
  cross-lingual representation learning at scale}.
\newblock In \emph{Proceedings of the 58th Annual Meeting of the Association
  for Computational Linguistics}, pages 8440--8451, Online. Association for
  Computational Linguistics.

\bibitem[{Conneau and Lample(2019)}]{Conneau2019}
Alexis Conneau and Guillaume Lample. 2019.
\newblock {Cross-lingual Language Model Pretraining}.
\newblock In \emph{Proceedings of the 2019 Conference on Advances in Neural
  Information Processing Systems (NeurIPS'19)}, Online. Curran Associates, Inc.

\bibitem[{Conneau et~al.(2018)Conneau, Rinott, Lample, Williams, Bowman,
  Schwenk, and Stoyanov}]{conneau-etal-2018-xnli}
Alexis Conneau, Ruty Rinott, Guillaume Lample, Adina Williams, Samuel Bowman,
  Holger Schwenk, and Veselin Stoyanov. 2018.
\newblock \href {https://doi.org/10.18653/v1/D18-1269} {{XNLI}: Evaluating
  cross-lingual sentence representations}.
\newblock In \emph{Proceedings of the 2018 Conference on Empirical Methods in
  Natural Language Processing}, pages 2475--2485, Brussels, Belgium.
  Association for Computational Linguistics.

\bibitem[{D'Amour et~al.()D'Amour, Heller, Moldovan, Adlam, Alipanahi, Beutel,
  Chen, Deaton, Eisenstein, Hoffman, Hormozdiari, Houlsby, Hou, Jerfel,
  Karthikesalingam, Lucic, Ma, McLean, Mincu, Mitani, Montanari, Nado,
  Natarajan, Nielson, Osborne, Raman, Ramasamy, Sayres, Schrouff, Seneviratne,
  Sequeira, Suresh, Veitch, Vladymyrov, Wang, Webster, Yadlowsky, Yun, Zhai,
  and Sculley}]{DAmour2020}
Alexander D'Amour, Katherine Heller, Dan Moldovan, Ben Adlam, Babak Alipanahi,
  Alex Beutel, Christina Chen, Jonathan Deaton, Jacob Eisenstein, Matthew~D.
  Hoffman, Farhad Hormozdiari, Neil Houlsby, Shaobo Hou, Ghassen Jerfel, Alan
  Karthikesalingam, Mario Lucic, Yian Ma, Cory McLean, Diana Mincu, Akinori
  Mitani, Andrea Montanari, Zachary Nado, Vivek Natarajan, Christopher Nielson,
  Thomas~F. Osborne, Rajiv Raman, Kim Ramasamy, Rory Sayres, Jessica Schrouff,
  Martin Seneviratne, Shannon Sequeira, Harini Suresh, Victor Veitch, Max
  Vladymyrov, Xuezhi Wang, Kellie Webster, Steve Yadlowsky, Taedong Yun,
  Xiaohua Zhai, and D.~Sculley.
\newblock \href {http://arxiv.org/abs/2011.03395} {Underspecification presents
  challenges for credibility in modern machine learning}.
\newblock \emph{CoRR}.

\bibitem[{Devlin et~al.(2019)Devlin, Chang, Lee, and
  Toutanova}]{devlin-etal-2019-bert}
Jacob Devlin, Ming-Wei Chang, Kenton Lee, and Kristina Toutanova. 2019.
\newblock \href {https://doi.org/10.18653/v1/N19-1423} {{BERT}: Pre-training of
  deep bidirectional transformers for language understanding}.
\newblock In \emph{Proceedings of the 2019 Conference of the North {A}merican
  Chapter of the Association for Computational Linguistics: Human Language
  Technologies, Volume 1 (Long and Short Papers)}, pages 4171--4186,
  Minneapolis, Minnesota. Association for Computational Linguistics.

\bibitem[{Dufter and Sch{\"u}tze(2020)}]{dufter-schutze-2020-identifying}
Philipp Dufter and Hinrich Sch{\"u}tze. 2020.
\newblock \href {https://doi.org/10.18653/v1/2020.emnlp-main.358} {Identifying
  elements essential for {BERT}{'}s multilinguality}.
\newblock In \emph{Proceedings of the 2020 Conference on Empirical Methods in
  Natural Language Processing (EMNLP)}, pages 4423--4437, Online. Association
  for Computational Linguistics.

\bibitem[{Ganin and Lempitsky(2015)}]{Ganin2015}
Yaroslav Ganin and Victor Lempitsky. 2015.
\newblock \href {http://proceedings.mlr.press/v37/ganin15.html} {Unsupervised
  domain adaptation by backpropagation}.
\newblock In \emph{Proceedings of the 32nd International Conference on Machine
  Learning}, volume~37 of \emph{Proceedings of Machine Learning Research},
  pages 1180--1189, Lille, France. PMLR.

\bibitem[{Gonen et~al.(2020)Gonen, Ravfogel, Elazar, and
  Goldberg}]{gonen-etal-2020-greek}
Hila Gonen, Shauli Ravfogel, Yanai Elazar, and Yoav Goldberg. 2020.
\newblock \href {https://doi.org/10.18653/v1/2020.blackboxnlp-1.5} {It{'}s not
  {G}reek to m{BERT}: Inducing word-level translations from multilingual
  {BERT}}.
\newblock In \emph{Proceedings of the Third BlackboxNLP Workshop on Analyzing
  and Interpreting Neural Networks for NLP}, pages 45--56, Online. Association
  for Computational Linguistics.

\bibitem[{Hu et~al.(2020)Hu, Ruder, Siddhant, Neubig, Firat, and
  Johnson}]{Hu2020}
Junjie Hu, Sebastian Ruder, Aditya Siddhant, Graham Neubig, Orhan Firat, and
  Melvin Johnson. 2020.
\newblock \href {http://arxiv.org/abs/2003.11080v5} {Xtreme: A massively
  multilingual multi-task benchmark for evaluating cross-lingual
  generalization}.
\newblock \emph{CoRR}.

\bibitem[{K et~al.(2020)K, Wang, Mayhew, and
  Roth}]{Karthikeyan2020Cross-Lingual}
Karthikeyan K, Zihan Wang, Stephen Mayhew, and Dan Roth. 2020.
\newblock \href {https://openreview.net/forum?id=HJeT3yrtDr} {Cross-lingual
  ability of multilingual bert: An empirical study}.
\newblock In \emph{International Conference on Learning Representations}.

\bibitem[{Lauscher et~al.(2020)Lauscher, Ravishankar, Vuli{\'c}, and
  Glava{\v{s}}}]{lauscher-etal-2020-zero}
Anne Lauscher, Vinit Ravishankar, Ivan Vuli{\'c}, and Goran Glava{\v{s}}. 2020.
\newblock \href {https://doi.org/10.18653/v1/2020.emnlp-main.363} {From zero to
  hero: {O}n the limitations of zero-shot language transfer with multilingual
  {T}ransformers}.
\newblock In \emph{Proceedings of the 2020 Conference on Empirical Methods in
  Natural Language Processing (EMNLP)}, pages 4483--4499, Online. Association
  for Computational Linguistics.

\bibitem[{Libovick{\'y} et~al.(2020)Libovick{\'y}, Rosa, and
  Fraser}]{libovicky-etal-2020-language}
Jind{\v{r}}ich Libovick{\'y}, Rudolf Rosa, and Alexander Fraser. 2020.
\newblock \href {https://doi.org/10.18653/v1/2020.findings-emnlp.150} {On the
  language neutrality of pre-trained multilingual representations}.
\newblock In \emph{Findings of the Association for Computational Linguistics:
  EMNLP 2020}, pages 1663--1674, Online. Association for Computational
  Linguistics.

\bibitem[{Marneffe et~al.(2020)Marneffe, Ginter, Goldberg, Hajič, Manning,
  McDonald, Nivre, Petrov, Pyysalo, Schuster, Silveira, Tsarfaty, Tyers, and
  Zeman}]{Marneffe2020}
Marie-Catherine~de Marneffe, Filip Ginter, Yoav Goldberg, Jan Hajič,
  Christopher Manning, Ryan McDonald, Joakim Nivre, Slav Petrov, Sampo Pyysalo,
  Sebastian Schuster, Natalia Silveira, Reut Tsarfaty, Francis Tyers, and Dan
  Zeman. 2020.
\newblock \href {https://universaldependencies.org/} {Universal dependencies
  v2.7}.

\bibitem[{Phang et~al.(2020)Phang, Calixto, Htut, Pruksachatkun, Liu, Vania,
  Kann, and Bowman}]{Phang2020}
Jason Phang, Iacer Calixto, Phu~Mon Htut, Yada Pruksachatkun, Haokun Liu, Clara
  Vania, Katharina Kann, and Samuel~R. Bowman. 2020.
\newblock \href {https://aclanthology.org/2020.aacl-main.56} {{E}nglish
  intermediate-task training improves zero-shot cross-lingual transfer too}.
\newblock In \emph{Proceedings of the 1st Conference of the Asia-Pacific
  Chapter of the Association for Computational Linguistics and the 10th
  International Joint Conference on Natural Language Processing}, pages
  557--575, Suzhou, China. Association for Computational Linguistics.

\bibitem[{Pires et~al.(2019)Pires, Schlinger, and
  Garrette}]{pires-etal-2019-multilingual}
Telmo Pires, Eva Schlinger, and Dan Garrette. 2019.
\newblock \href {https://doi.org/10.18653/v1/P19-1493} {How multilingual is
  multilingual {BERT}?}
\newblock In \emph{Proceedings of the 57th Annual Meeting of the Association
  for Computational Linguistics}, pages 4996--5001, Florence, Italy.
  Association for Computational Linguistics.

\bibitem[{Rama et~al.(2020)Rama, Beinborn, and Eger}]{rama-etal-2020-probing}
Taraka Rama, Lisa Beinborn, and Steffen Eger. 2020.
\newblock \href {https://doi.org/10.18653/v1/2020.coling-main.105} {Probing
  multilingual {BERT} for genetic and typological signals}.
\newblock In \emph{Proceedings of the 28th International Conference on
  Computational Linguistics}, pages 1214--1228, Barcelona, Spain (Online).
  International Committee on Computational Linguistics.

\bibitem[{Rosenberg and Hirschberg(2007)}]{Rosenberg2007}
Andrew Rosenberg and Julia Hirschberg. 2007.
\newblock \href {https://aclanthology.org/D07-1043} {{V}-measure: A conditional
  entropy-based external cluster evaluation measure}.
\newblock In \emph{Proceedings of the 2007 Joint Conference on Empirical
  Methods in Natural Language Processing and Computational Natural Language
  Learning ({EMNLP}-{C}o{NLL})}, pages 410--420, Prague, Czech Republic.
  Association for Computational Linguistics.

\bibitem[{Ruder et~al.(2017)Ruder, Vuli{\'{c}}, and S{\o}gaard}]{Ruder2017}
Sebastian Ruder, Ivan Vuli{\'{c}}, and Anders S{\o}gaard. 2017.
\newblock \href {https://doi.org/10.1177/0964663912467814} {{A Survey Of
  Cross-lingual Word Embedding Models}}.
\newblock \emph{CoRR}, pages 1--55.

\bibitem[{Sammons(2015)}]{Sammons2015}
Mark Sammons. 2015.
\newblock \href {https://doi.org/10.1162/COLI} {{Recognizing textual
  entailment}}.
\newblock In Chris {Lappin, Shalom and Fox}, editor, \emph{The Handbook of
  Contemporary Semantic Theory}, 2 edition, chapter~17, pages 523--557. John
  Wiley \& Sons Ltd.

\bibitem[{Tan and Joty(2021)}]{Tan2021}
Samson Tan and Shafiq Joty. 2021.
\newblock \href {https://doi.org/10.18653/v1/2021.calcs-1.19} {Code-mixing on
  sesame street: Dawn of the adversarial polyglots}.
\newblock In \emph{Proceedings of the Fifth Workshop on Computational
  Approaches to Linguistic Code-Switching}, page 141, Online. Association for
  Computational Linguistics.

\bibitem[{Vaswani et~al.(2017)Vaswani, Shazeer, Parmar, Uszkoreit, Jones,
  Gomez, and Kaiser}]{Vaswani2017}
Ashish Vaswani, Noam Shazeer, Niki Parmar, Jakob Uszkoreit, Llion Jones,
  Aidan~N. Gomez, and {\L}ukasz Kaiser. 2017.
\newblock \href {http://arxiv.org/abs/arXiv:1706.03762v5} {{Attention Is All
  You Need}}.
\newblock In \emph{Proceedings of the 31st Conference on Neural Informaton
  Processing Systems (NeurIPS'17)}, Long Beach, CA.

\bibitem[{Williams et~al.(2018)Williams, Nangia, and
  Bowman}]{williams-etal-2018-broad}
Adina Williams, Nikita Nangia, and Samuel Bowman. 2018.
\newblock \href {https://doi.org/10.18653/v1/N18-1101} {A broad-coverage
  challenge corpus for sentence understanding through inference}.
\newblock In \emph{Proceedings of the 2018 Conference of the North {A}merican
  Chapter of the Association for Computational Linguistics: Human Language
  Technologies, Volume 1 (Long Papers)}, pages 1112--1122, New Orleans,
  Louisiana. Association for Computational Linguistics.

\bibitem[{Wu and Dredze(2019)}]{wu-dredze-2019-beto}
Shijie Wu and Mark Dredze. 2019.
\newblock \href {https://doi.org/10.18653/v1/D19-1077} {Beto, bentz, becas: The
  surprising cross-lingual effectiveness of {BERT}}.
\newblock In \emph{Proceedings of the 2019 Conference on Empirical Methods in
  Natural Language Processing and the 9th International Joint Conference on
  Natural Language Processing (EMNLP-IJCNLP)}, pages 833--844, Hong Kong,
  China. Association for Computational Linguistics.

\end{thebibliography}
